\documentclass{article}

\usepackage{arxiv}

\usepackage{booktabs}
\usepackage[utf8]{inputenc} % allow utf-8 input
\usepackage[T1]{fontenc}    % use 8-bit T1 fonts
\usepackage{hyperref}       % hyperlinks
\usepackage{url}            % simple URL typesetting
\usepackage{booktabs}       % professional-quality tables
\usepackage{amsfonts}       % blackboard math symbols
\usepackage{amsmath}
\usepackage{nicefrac}       % compact symbols for 1/2, etc.
\usepackage{microtype}      % microtypography
\usepackage{lipsum}		% Can be removed after putting your text content
\usepackage{graphicx}
\usepackage{tabularx}
\usepackage{doi}
\usepackage{subfig}
\usepackage{todonotes}
\usepackage{tablefootnote}
\usepackage{color}                 % colors
\definecolor{lightgrey}{rgb}{0.9,0.9,0.9} % color values Red, Green, Blue
\definecolor{lightred}{rgb}{0.9,0.7,0.7} % color values Red, Green, Blue
\definecolor{lightgreen}{rgb}{0.7,0.9,0.7} % color values Red, Green, Blue
\definecolor{lightblue}{rgb}{0.5,0.7,0.9} % color values Red, Green, Blue
\definecolor{blue}{rgb}{0.0,0.43,0.72}
\definecolor{red}{rgb}{0.6,0.0,0.0}
\definecolor{purple}{rgb}{0.63,0.13,0.94}

\newif\ifECCV
\newif\ifCifar
\newif\ifArchive
\Archivetrue

\newif\ifArchiveEx  %extra stuff originally placed for archive 
\newif\ifSupplement
\newcommand{\tabcapwidth}{1mm}

\title{CoShNet: A Hybrid Complex Valued Neural Network using Shearlets}

\author{ \href{https://orcid.org/0000-0001-9117-3772}{\includegraphics[scale=0.06]{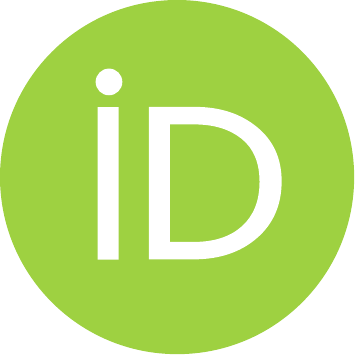}\hspace{1mm}Manny ~Ko} \\
	Independent researcher \\
	\texttt{man961@yahoo.com} \\
	\And
	\href{https://orcid.org/0000-0003-4903-5581}{\includegraphics[scale=0.06]{orcid.pdf}\hspace{1mm}Ujjawal K.~Panchal} \\
	Independent researcher \\
	\texttt{ujjawalpanchal32@gmail.com} \\
	 \And
	\href{https://orcid.org/0000-0000-0000-0000}{\includegraphics[scale=0.06]{orcid.pdf}\hspace{1mm}Héctor Andrade-Loarca} \\
	Mathematisches Institut der LMU München\\
	\texttt{arsenal997@hotmail.com} \\
	\And
	\href{https://orcid.org/0000-0000-0000-0000}{\includegraphics[scale=0.06]{orcid.pdf}\hspace{1mm}Andres Mendez-Vazquez} \\
	Computer Science, Cinvestav Guadalajara\\
	\texttt{andres.mendez@cinvestav.mx} \\
}

\renewcommand{\shorttitle}{\textit{arXiv} Template}

\hypersetup{
pdftitle={A template for the arxiv style},
pdfsubject={q-bio.NC, q-bio.QM},
pdfauthor={David S.~Hippocampus, Elias D.~Striatum},
pdfkeywords={First keyword, Second keyword, More},
}

\begin{document}
\maketitle

\begin{abstract}
	In a hybrid neural network, the expensive convolutional layers are replaced by a non-trainable fixed transform with a great reduction in parameters. 
    In previous works, good results were obtained by replacing the convolutions with wavelets.
	However, wavelet based hybrid network inherited wavelet's lack of vanishing
	moments along curves and has axis-bias. We propose to use Shearlets with its 
	robust support for important image features like edges, ridges and blobs. The resulting network is called
	Complex Shearlets Network (CoShNet). It was tested on Fashion-MNIST against ResNet-50 and
	Resnet-18, obtaining $92.2\%$ versus $90.7\%$ and $91.8\%$ respectively. 
	The proposed network has $49.9k$ parameters versus ResNet-18 with $11.18m$ 
    and use $52\times$ fewer FLOPs.   
 Finally, we trained in under 20 epochs versus 200 epochs required by ResNet and do not
	need any hyperparameter tuning nor regularization.

Code: \url{https://github.com/Ujjawal-K-Panchal/coshnet}
\end{abstract}


%
% Introduction
% 
\section{Introduction}
Neural networks have been getting larger each day, leading
to an explosion in required training time and large amount of quality labeled data. 
Often, this becomes the main limiting
factor to successfully deploying a deep neural network. Training such large networks has
adverse environmental effects and requires GPU clusters with large memory that are out 
of reach for a lot of organizations. 
The authors feel standard CNNs unnecessarily spend a lot of model capacity and computational resources to
learn a good embedding (features) in order to feed the classifier layers. In addition, they need
extensive hyperparamter tuning to land the ``lottery ticket'' \cite{JonathanFrankle2019}
which exacerbates the heavy computation burden. 

The best known hybrid neural network is Oyallon \& Mallat's scattering-network \cite{Oyallon2015}. It replaces the expensive convolution layers with a \textit{fixed but well designed signal transform} (wavelets)
that produces a sparse embedding. 
Mallat \cite{Bruna2012} also eloquently showed that scattering give us the highly desirable
property of \textit{local deformation stability} and \textit{rotational invariance}.
In \cite{Szegedy2013} Szegedy et al. documented the fragility of standard CNNs and their vulnerability to input perturbations.
However, scattering networks can be further improved with a better transform inside a complex-valued network.

\begin{figure}[htbp]
	\centering
	\begin{tabular}{ccc}
		\subfloat[Source Lena image.]{\includegraphics[width = 0.201\linewidth]{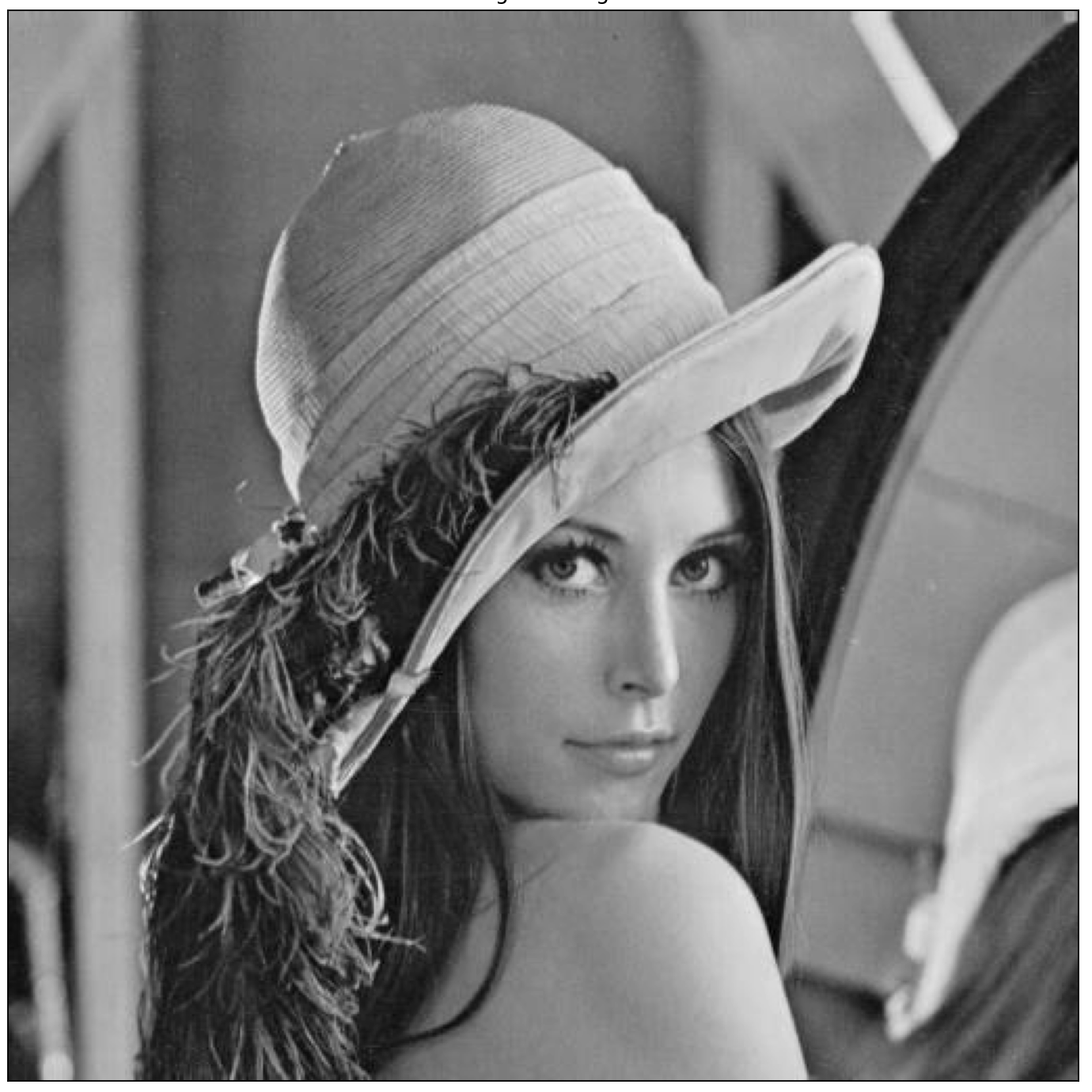}} &
		\subfloat[Edge detection using Haar wavelet transform]{\includegraphics[width = 0.199\linewidth]{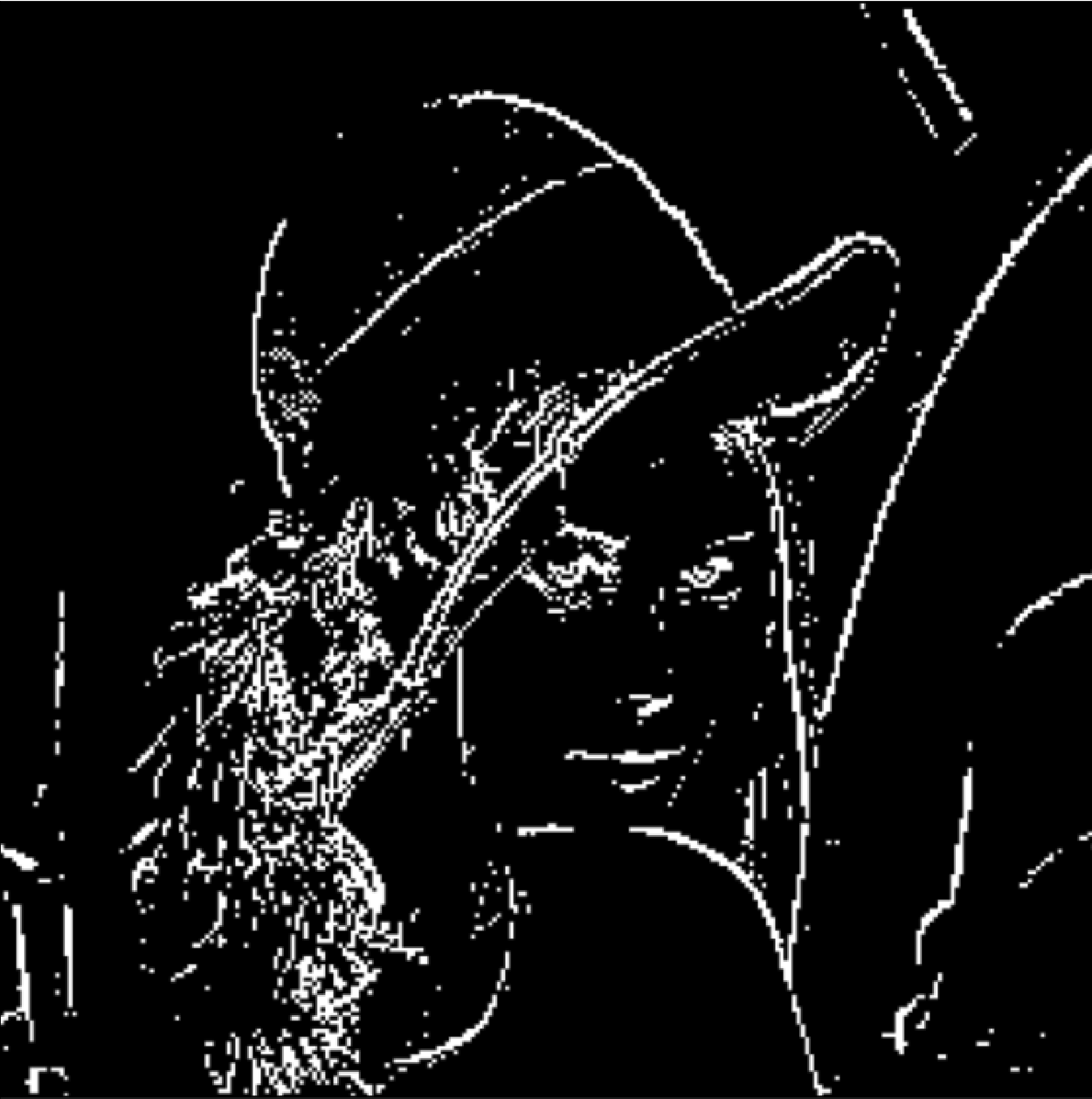}} &
		\subfloat[Edge detection using CoShRem transform]{\includegraphics[width = 0.2\linewidth]{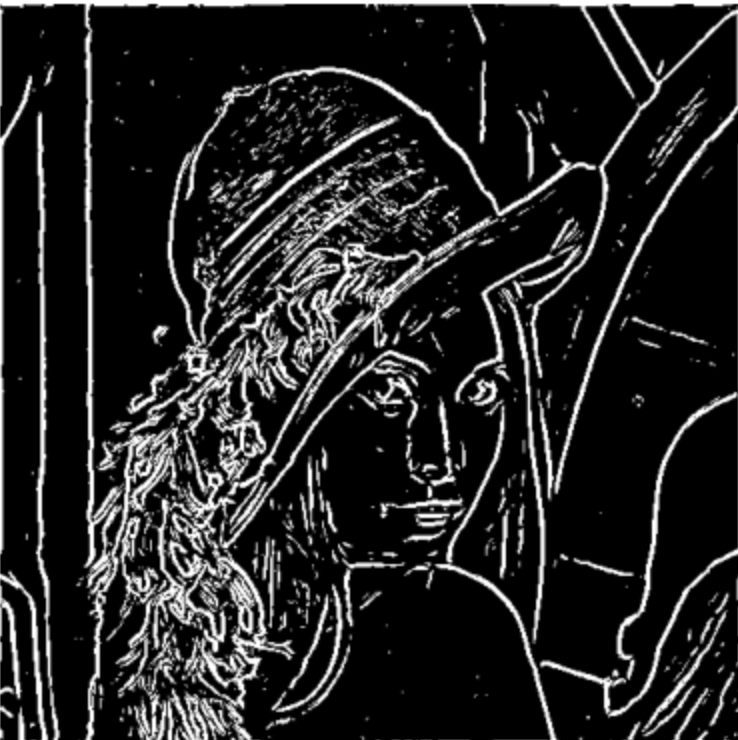}}\\
	\end{tabular}
	%\subfloat[Edge detection using Coiflet Wavelets]{\includegraphics[width = 0.15\linewidth]{./images/Lena/Lena-CoifletWavelet.png}} \hspace{2mm}
	%\subfloat[Edge detection using CoShREM]{\includegraphics[width = 0.15\linewidth]{./images/Lena/Lena-CoShREM.png}} \hspace{2mm}
	\caption{
		wavelet transform fails to capture curvilinear edges which are successfully 
		captured by the CoShRem.
	}
	\label{fig:Lena}
\end{figure}

These are the main reasons behind CoShNet's novel architecture.
CoShNet uses a complex Shearlet transform
\href{http://www.math.uni-bremen.de/cda/software.html}{\textbf{CoShRem}} \cite{Kutyniok2014c} \cite{Reisenhofer2019} and operates entirely in $\mathbb{C}$-domain. 
CoShRem does not have axis bias which wavelets suffers from, and handles edges of all orientations. Fig. \ref{fig:Lena} contrasts wavelets' and shearlets' ability to
represent edges. CoShRem gracefully handles edges of varies orientations while wavelets
struggles with some of them (curved or non-aligned to horizontal or vertical axes).
CoShRem also gives us a critical property call ``phase-congruency'' \cite{Kovesi1999}, 
which keeps the features stable under perturbation and noise.
Kovesi eloquently uses the agreement of phase across scale (frequency)
to robustly localize an edge/ridge/blob while the gradients and magnitude can disagree
extensively (fig \ref{fig:phasecong-f3}). 
\ifArchive Phase-congruency will be covered in detail in S \ref{phase}. \fi
This is in contrast to scattering
and it's use of complex-modulus which kills the critically important phase information. 
Recently, Wiatowski \cite{Wiatowski2015} shows us, CoShRem is a type of scattering
and inherits its desirable property of deformation stability and can be efficiently
implemented by FFT. Besides using a robust transform, CoShNet has the following desirable
properties:

\ifArchive \section{Contributions} \fi

\begin{itemize}
%bullet 1
\item \textbf{A Compact hybrid and efficient CNN architecture}.  CoShNet uses a fixed 2nd-generation wavelet transform
to produce informative and stable representations for classification.
In addition, through a series of tests, CoShNet has demonstrated state of the art results using a fraction of parameters ($49.9k$) versus 
ResNet-18 with $11.2m$. 
The architecture does not require any regularization nor hyperparameter tuning and is extremely stable with respect to initialization and training.

%bullet 2
\item \textbf{Exceptional Generalization}. Training with the $10k$ samples
from Fashion and testing using the remaining $60k$ samples, 
we are able to obtain Top-1 accuracy close to $90\%$.

\ifArchive  %Andres feel this is impl. detail and does not belong here
\item \textbf{Complex-valued neural network} (CVnn)
a CVnn has many attractive properties absent in its real valued counterpart. 
Nitta \cite{Nitta2002}\cite{Nitta2013} studied the critical-points of real vs. complex neural networks and showed that a lot of critical-points of regular CNNs
are local minima, whereas CVnn's critical-points are mostly saddle-points.
Nitta \cite{Nitta2003} also showed that the decision surface \ref{cplx-decision} of a CVnn has more representational power and is self regularizing.
However, a good implementation in the complex domain is non-trivial.
In S. \ref{cvnn} we will layout the important components for a efficient CVnn that performs
competitively.
\fi

%bullet 3
\item \textbf{Fully complex CVnn} CoShRem is unique among CVnns as a fully complex network - all operations are performed using complex operators.

%bullet 4
\item \textbf{Training a compressed model}. We present a novel way
of training a highly compressed version of CoShNet by using tensor compression and DCFNet \cite{Qiu2018}
while preserving all of its good qualities.
Unlike most efforts on model compression, the compressed model is trained in one pass.
\end{itemize}
Extensive experimental data presented here demonstrates that CoShNet trains much faster and uses less training data. 
All of the proposed models can be trained in less than $20$ epochs 
(some as low as $10$) in under 20 seconds on a modest GTX 1070Ti.
\ifArchive Contrast that with a recent \href{https://colab.research.google.com/github/shoji9x9/Fashion-MNIST-By-ResNet/blob/master/Fashion-MNIST-by-ResNet-50.ipynb#scrollTo=iQoh1m9oit4-}{ResNet-50 based model}
which took 400 epochs to reach $91\%$ and has 28m parameters.

Finally, the structure of the paper is as follow:
Section \ref{hybrid-nn} discusses some critical properties of complex Shearlets
and the importance of phase-congruency.
Section \ref{cvnn} outlines the mathematics and implementation of complex-valued neural network (CVnn).
Section \ref{coshcvnn} presents the simple architecture of \textbf{CoShNet}.
Section \ref{compress} presents the highly compressed tiny-\textbf{CoShNet}.
Section \ref{experiments} presents the experimental results and ablation studies.

%
% Hybrid NN.
%
\section{Hybrid CNN \& Scattering} \label{Bruna2012}\label{hybrid-nn}
Raghu et al. showed in \cite{raghu2017svcca} that the lower levels of a CNN converge very early
and that \textit{class-specific information is concentrated in the upper layers}.
Their proposal is to freeze the lower levels early.
This confirms the central thesis of CoShNet \footnote{We were pursuing the current line of investigation and came upon \cite{raghu2017svcca} only recently.}
for the role played by the lower levels in a classic CNN
which is to learn an embedding. 
This article argues \textit{the embedding obtained by the CNN is largely generic} -
it has more to do with general ``image statistics'' and less to do with class-specific traits.
Ravishankar et al. \cite{Ravishankar2018} ``Deep Residual Transform'' produce
learned filters \ref{fig:ravi-f1} which look very similar to those of K-SVD
\ref{fig:ksvd-f1}, 
which is an exemplar of sparsity methods.
These results inspired us to go one step further, and use a fixed signal transform 
instead of using a lot of capacity and time to learn a sparsifying transform.
A fixed transform automatically implies better sampling efficiency and less affected
by distribution or covariant shift. 

\begin{figure}
\ifECCV \vspace{-6mm} \fi
\centering
\subfloat[Learned Transforms \cite{Ravishankar2018} \label{fig:ravi-f1}]{
	\includegraphics[width=0.195\linewidth]{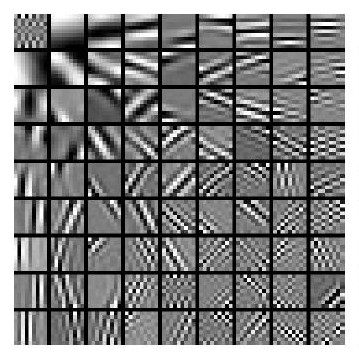}
}
\qquad
\subfloat[Atoms from 
\href{https://www.researchgate.net/publication/322024484_Orthogonal_Matching_Pursuit_and_K-SVD_for_Sparse_Encoding}{K-SVD for Sparse Encoding} \label{fig:ksvd-f1}]
{
	\includegraphics[width=0.6\linewidth]{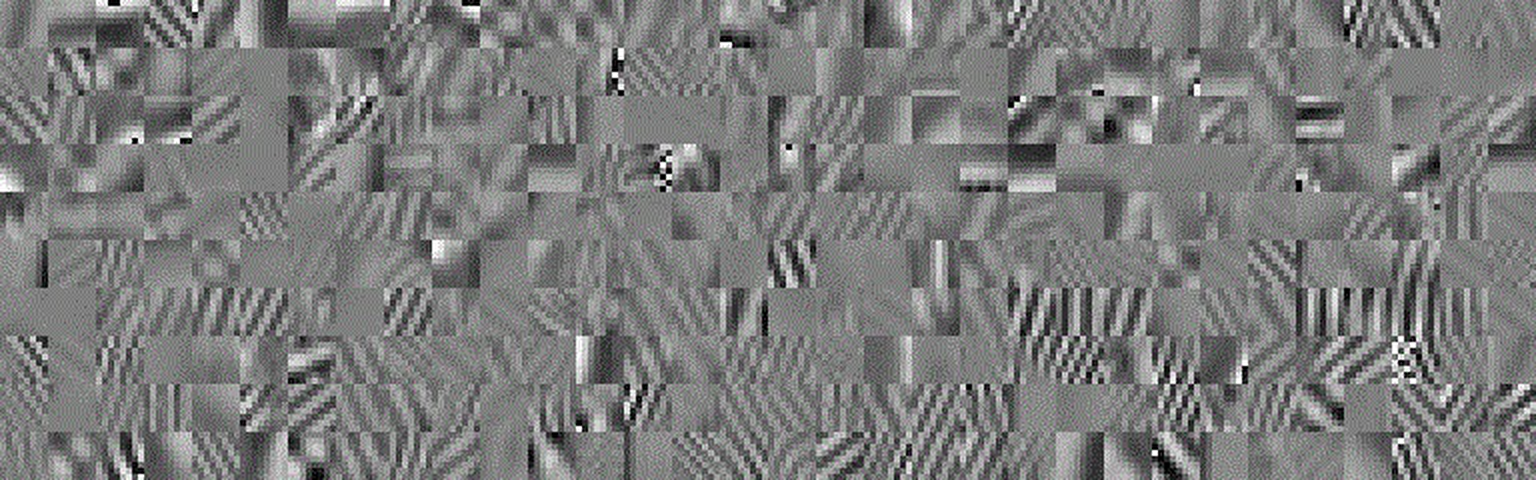}
}
\ifECCV \vspace{-2mm} \fi
\caption{a learned transform from \cite{Ravishankar2018} and atoms in a sparse encoding}
\ifECCV \vspace{-5mm} \fi
\end{figure}

\label{Oyallon2015}
Thus, CoShNet can be evaluated in the context of Oyallon \& Mallat's scattering-network
\cite{Oyallon2015}. Both are hybrid NN, and they replace the early layers in a 
CNN with a 
\textit{fixed transform} that provide a sparse embedding for images. 
Mallat \cite{Bruna2012} also eloquently showed that scattering gave us the highly desirable
property of \textit{local deformation stability
\footnote{See \cite{Szegedy2013}
for the fragility of CNNs and their vulnerability to input perturbations.}
and rotation invariance}.
\ifSupplement \footnote{More details on scattering are in \ref{cplx-scattering-supp} of the supplement}. \fi 
However, scattering use complex-modulus after convolving with a Morlet wavelet 
\ifArchive \ref{fig:scatter-f4}  \fi
which kills the phase information. Phase encodes structure and is rich in semantics.
CoShRem carefully preserves the phase across the cascade
of scale and orientation by performing all operation in $\mathbb{C}$. 
We will discuss the critical role of phase in S 2.2.

\ifArchive
\begin{figure}
\centering
\includegraphics[width=0.6\linewidth]{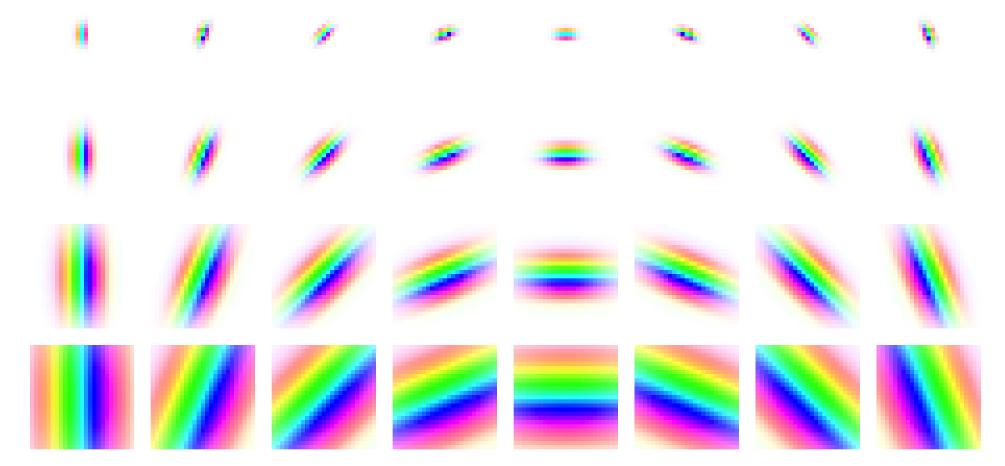}
\caption{Real and imaginary parts of Morlet wavelets at 4 different scales and 8 orientations. Phase and amplitude are respectively given by the color and the contrast.
Scattering and CoShRem both using filters that resemble Morlet wavelets.
}
\label{fig:scatter-f4}
\end{figure}
\fi

%
% CoShREM Transform
% 
\ifArchive \subsection{CoShRem Transform} \fi
%
% Note: Use onsen/visuals/new-visuals/slides-viz.ipynb to recreate the figures of your interest from this subsection.
%

%
% https://github.com/dedale-fet/alpha-transform
%

CoShRem is based on 
$\alpha$-molecules which is introduced in \cite{Grohs2013}.
The transform is constructed by translating, scaling, and rotating members of a possibly infinite set of generator functions $\{\psi\}$.
The scaling $a$ and shearing $s$ are applied to the arguments of the generator:
\begin{align}
&A_{a, \alpha} S_{s} \psi(x - t),\\
&A_{a, \alpha} = \begin{bmatrix}
  a & 0 \\ 0 & a^{\alpha}
\end{bmatrix},
\hspace{4pt} S_s = \begin{bmatrix}1 & s\\
0 & 1
\end{bmatrix},
 \hspace{4pt} a > 0, \hspace{2pt} \alpha \in [0,1], \nonumber
\end{align}

\noindent
$\alpha$ controls the anisotropy and smoothly interpolates the generator function to create a family of
wavelets (isotropic scaling with $\alpha = 1$),
shearlets (parabolic scaling with $\alpha = 0.5$) and ridgelets
(fully anisotropic scaling with $\alpha = 0$).

\textbf{CoShRem} can stably detects edges, ridges, and blobs
\cite{Reisenhofer2019} using phase and oriented filter-banks \ref{fig:phasecong-f3}
without the need for unstable gradients. 
In Table \ref{tab:featXmol}, one can clearly see the filter's alignment 
and the image features it detects.
As a Fourier integral operator, \textbf{CoShRem} is stable and noise immune \cite{Guo2008}.

%
% CoShREM maximal response - MNIST
%
\begin{table}
	\centering
	\begin{tabular}{|c|c|c|c|c|}
		\hline image / filter &
			   \includegraphics[width = 12mm]{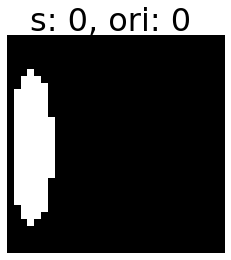} &
			   \includegraphics[width = 12mm]{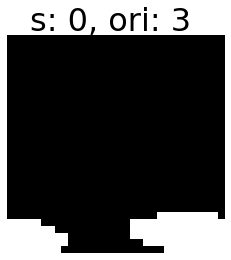} &
			   \includegraphics[width = 12mm]{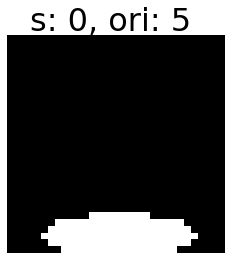} &
			   \includegraphics[width = 12mm]{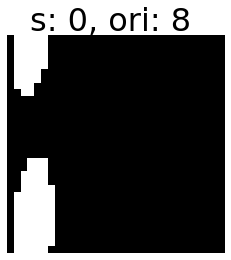} \\
		
		\hline \includegraphics[width = 12mm]{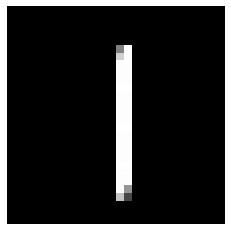}
				& $11.13\%$ & $1.03\%$ & $1.04\%$ & $10.17\%$\\
		\hline \includegraphics[width = 12mm]{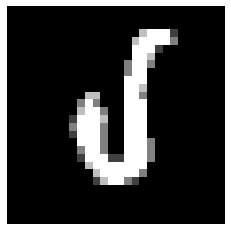}
				& $7.96\%$ & $3.83\%$ & $3.96\%$ & $7.42\%$\\
		\hline \includegraphics[width = 12mm]{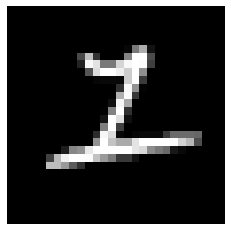}
				& $4.43\%$ & $7.35\%$ & $7.73\%$ & $4.16\%$\\
		\hline \includegraphics[width = 12mm]{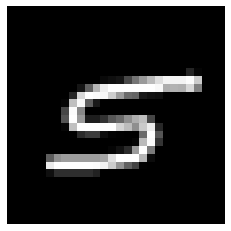}
				& $2.90\%$ & $8.79\%$ & $9.33\%$ & $2.78\%$\\
		\hline
	\end{tabular}
	\vspace{1mm}
	\caption{columns: 4 CoShRem filters, rows: MNIST. Each cell is the percentage energy of the filter response. In the first two rows, more energy is concentrated in the vertical aligned filters since both are dominated by vertical shapes and the opposite for remain two rows.}
	\label{tab:featXmol}
\end{table}

Reisenhofer at al. \cite{Reisenhofer2019} show that the CoShRem transform is very
stable in the presence of perturbations.
Fig \ref{fig:perturbed-coshrem} shows despite the considerable perturbations
(blurring and Gaussian noise),
CoShRem remain stable to most of the characteristic 
edges and ridges (two step discontinuity in close proximity).

\begin{figure}[htbp]
\ifECCV \vspace{-5mm} \fi
	\centering
	\subfloat[image, noise($\sigma^2 = 50$), blur($\sigma = 1$)]{
		{\includegraphics[width = 24mm]{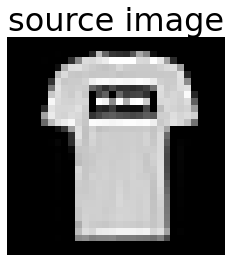}}
		{\includegraphics[width = 24mm]{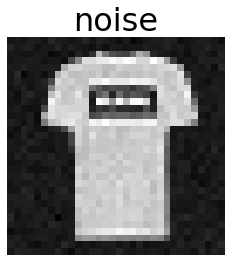}}
		{\includegraphics[width = 24mm]{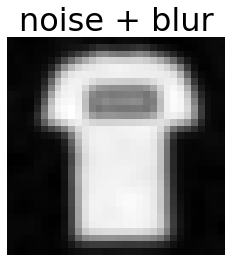}}
	}\\
	\subfloat[phase without perturation]{
		{\includegraphics[width = 18mm]{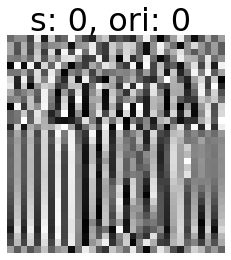}}
		{\includegraphics[width = 18mm]{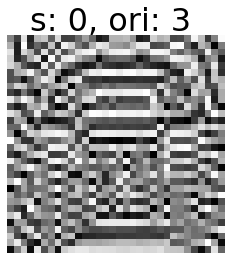}}
		{\includegraphics[width = 18mm]{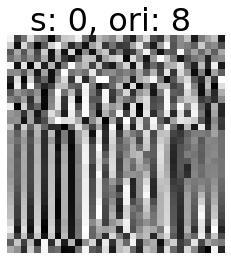}}
	}\hspace{1mm}
	\subfloat[magnitude without perturbation]{
		{\includegraphics[width = 18mm]{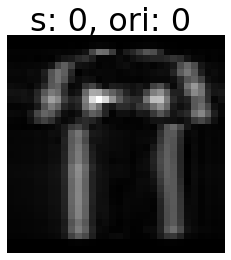}}
		{\includegraphics[width = 18mm]{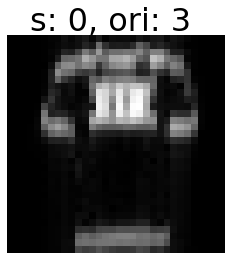}}
		{\includegraphics[width = 18mm]{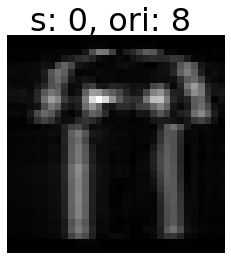}}
	}\\
	\subfloat[phase with perturbation]{
		{\includegraphics[width = 18mm]{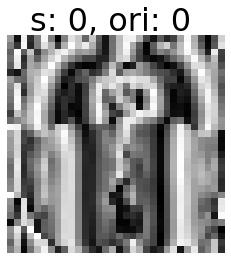}}
		{\includegraphics[width = 18mm]{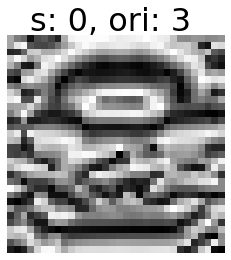}}
		{\includegraphics[width = 18mm]{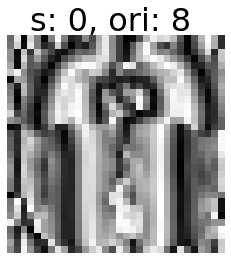}}
	}\hspace{1mm}
	\subfloat[magnitude with perturbation]{
		{\includegraphics[width = 18mm]{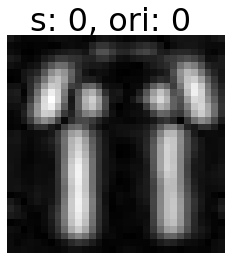}}
		{\includegraphics[width = 18mm]{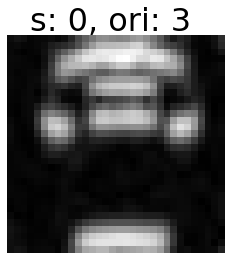}}
		{\includegraphics[width = 18mm]{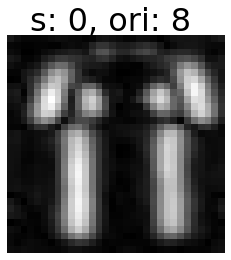}}
	}
	\caption{(a) image and perturbations (b)..(e) CoShRem phase, magnitude}
	\label{fig:perturbed-coshrem}
\end{figure}

%
% Phase & Phase-congruency
% 
\ifArchive \subsection{Phase \& Phase-congruency}\label{phase} \fi
The important role of \textit{phase} has been well established in signal processing.
A motivating example is \cite{Huang1975}, \cite{Oppenheim1981} for encoding structure
and semantics. 
Phase play a critical role in Kovesi \cite{Kovesi1999}'s phase-congruency". 
He clearly show the power of agreement of phase across scales and orientation\ifArchive \cite{Kovesi2003}\fi. 
Reisenhofer et al. \cite{Reisenhofer2019} and \cite{Andrade-Loarca2020} show that
with ``phase-congruency", CoShRem can extract
stable features - edges, ridges and blobs  - that are contrast invariant.
In Fig \ref{fig:phasecong-f1}.b we can see a stable and robust
(immune to noise and contrast variations) 
localization of critical features in an image by using agreement of phase. 
This is all the more striking in contrast to Fig \ref{fig:phasecong-f1}  
where gradients across scales wildly mismatch and are extremely sensitive to noise.
In other words, gradients fluctuates wildly across scale but phase remains very
stable at critical parts of the image. 

%\textbf{CoShRem} is fully idempotent
%\footnote{idempotent: applying the feature detector to its output and getting back the same set %of features} and phase congruent. 

\ifSupplement
\footnote{See \ref{Hilbert} in the supplement for more details.}
\fi
\begin{figure}
\ifECCV \vspace{-4mm} \fi
	\centering
    \includegraphics[width=0.22\linewidth]{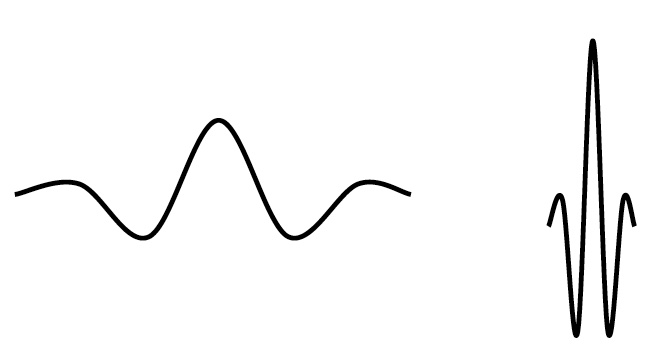}
    \hspace{50pt}
	\includegraphics[width=0.22\linewidth]{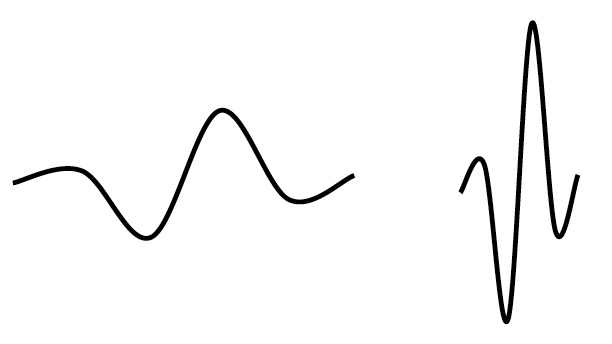}
	\caption[]{(a) even-symmetric Gabor (b) odd-symmetric Gabor}
	\label{fig:phasecong-f3}
\ifECCV \vspace{-5mm} \fi
\end{figure}

\label{Hilbert}
Phase congruency \cite{Morrone1988} in CoShRem is implemented using even-odd complex
symmetric
\footnote{a function is even-symmetric if $f(x) = f(-x)$, odd-symmetric if $f(x) = -f(-x)$}
Gabor-pairs
- filters of equal amplitude
but \textit{orthogonal in phase} - i.e. a $\pi/2$ offset (Fig. \ref{fig:phasecong-f3}).
\ifArchive
See  \href{https://ccia.ugr.es/cvg/REWIC/articulo/frewic/html/chapter1/node1.html}{Phase congruency} for more details.
\fi
The pair is related by a Hilbert transform, which is the textbook method
from Bracewell \cite{bracewell1978fourier} to convert a real signal to a complex one. CoShNet thus inherits
the desirable properties of the Hilbert transform.
%\todo{Write up other CVNN's struggle in this area of converting images to complex.}

\begin{figure}
\ifECCV \vspace{-3mm} \fi
\centering
\includegraphics[width=0.65\linewidth]{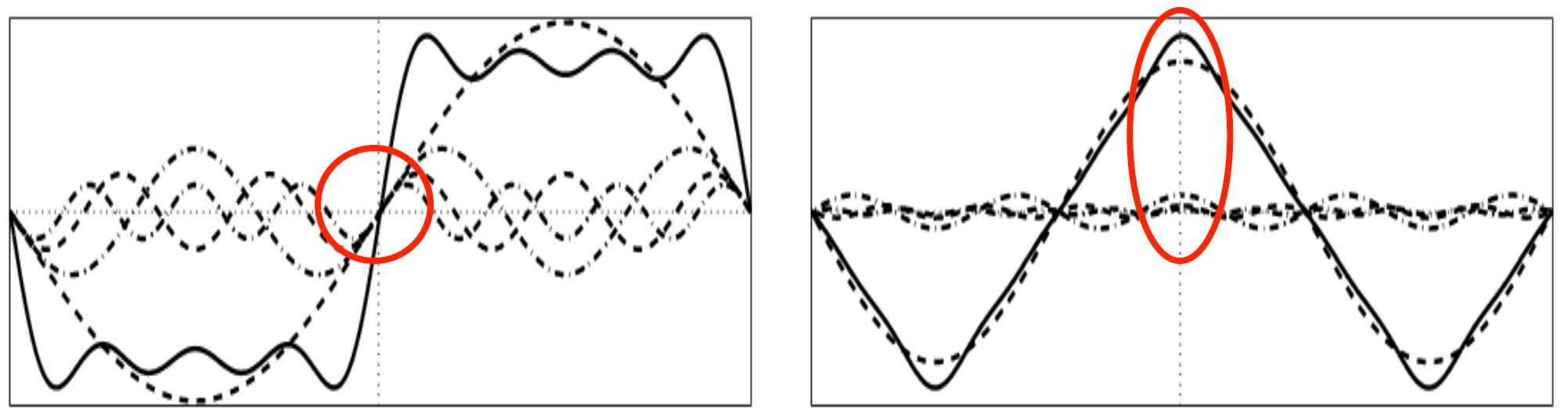}
\caption[]{phase-congruency - (a) gradients across scales disagree (b) Fourier components are all in phase at the step in the
square wave and at the peaks and troughs of the triangular wave. Images generated using \cite{KovesiMATLABCode}.}
\label{fig:phasecong-f1}
\ifECCV \vspace{-1mm} \fi
\end{figure}

%
% Complex-valued CNN
% 
\ifECCV \vspace{-8mm} \fi
\section{Complex-Valued Neural Networks (CVnns)} \label{cvnn}
Given the importance of phase and CoShRem being a complex transform, it seems
natural for the network to be complex-valued. A CVnn has complex 
inputs, weights, biases and activations with all the mathematical operation performed in $\mathbb{C}$.
It requires a complex version of \textsf{backprop} \cite{Benvenuto92} demanding
careful analysis and efficient implementation \ref{cplx-backprop}.
It can be traced back to the pioneering work of
Nitta \cite{Nitta1997} and Hirose \cite{Hirose2011}.
It has several attractive properties such as orthogonal decision boundaries and 
attractive structure among its critical points, which are particularly powerful, but not well known.

\vspace{-2mm}
\subsection{Complex Decision Boundary, Critical Points \& Generalization}\label{cplx-decision}
Nitta's \cite{Nitta2002}, \cite{Nitta2003} seminal work shows the 
CVnn has the following remarkable properties not found in standard CNNs -
\begin{figure}[htbp]
\vspace{-2mm}
\centering
\includegraphics[width=0.50\linewidth]{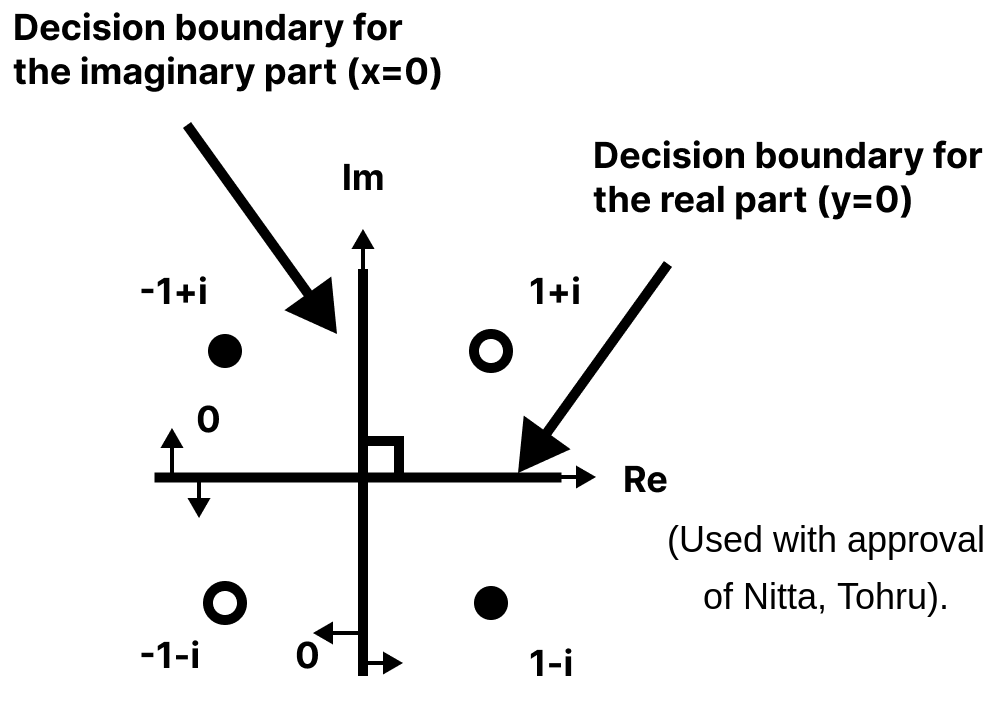}
\caption{Decision boundary formed by a neuron of a CVnn.}
\label{fig:nitta-f1}
\vspace{-6mm}
\end{figure}

%
% Nitta's 3 bullet points on CVnn
%
\begin{enumerate}
%bullet 1
\item \textbf{Orthogonality of decision boundaries}\label{Orthogonality}:
The \textit{decision boundary of a CVnn consists of two hypersurfaces that intersect
orthogonally} (Fig. \ref{fig:nitta-f1}) 
and \textit{divides a decision region into four equal sections}.
Furthermore, the decision boundary of a 3-layer CVnn stays almost orthogonal \cite{Nitta2003291}.
This orthogonality improves generalization.
As an example, several problems (e.g. Xor) that cannot be solved with a single real neuron,
can be solved with a single complex-valued neuron using the orthogonal property.

%bullet 2
\item \textbf{Structure of critical points}\label{critical-pts}:
Nitta shows most of the critical points of a CVnn are \textit{saddle points},
not local minima \cite{Nitta2013}, unlike the real-valued case \cite{Fukumizu2000}.
The derivative of the loss function is equal to zero at a critical point. 
SGD with random inits can largely avoid saddle points 
\cite{lee2016gradient} \cite{Jin2017}, but not a local minimum. 

%bullet 3
\item \textbf{Generalization}: a CVnn trains several times faster 
than regular CNN and generalizes much better. 
Nitta \cite{Nitta2003291} show they are both connected to Orthogonality property.
We will show in our experimental section how our CVnn can be trained with $1k$ data
and 20 epochs with excellent results.
\end{enumerate}
Our finding is in strong agreement of with all three of Nitta's findings.
The power of a NN is determined by its ability to model
complex decision boundaries \cite{shi2020}. A CVnn is endowed with a
much more powerful ability to represent complex decision surfaces due to
Orthogonality property \ref{Orthogonality}.

\subsection{Complex Product \& $2\times 2$ Orthogonal Matrix} \label{cplx-mul}
In neural networks, most of the heavy lifting are performed by linear-operators
%with \texttt{conv} being equivalent to a Toeplitz $W$
$\label{linear-xform}z = W\times x + b$.
For the $\mathbb{C}$ valued case, the complex-linear (\textsf{ear}) operator becomes:
\begin{equation}\label{cplx-mul2}
	z = \mathbb{R}(W) \times \mathbb{R}(x) - \mathbb{I}(W) \times \mathbb{I}(x) + i( \mathbb{I}(W) \times \mathbb{R}(x) + \mathbb{R}(W) 
	\times \mathbb{I}(x)) + b
\end{equation}

In terms of operator count, a \textsf{cplx-linear} is 4 times more expensive than a real one. We will later show the actual performance difference is considerably less
with a good implementation.
The \textsf{cplx-linear} used in CoShNet follows 
Hirose \cite{Hirose2009} and treats complex as a $2\times2$ orthogonal matrix.
Given $\mathbb{C}$-linear transform $T_c: c=a + ib$:
\begin{eqnarray}\label{eqn-3}
T_c\begin{pmatrix} x \\ y\end{pmatrix} = \begin{pmatrix} ax -by \\ bx + ay\end{pmatrix} 
= \begin{pmatrix} a & -b \\ b & a\end{pmatrix}\begin{pmatrix} x \\ y\end{pmatrix} \label{matrix-cplx-lin-xform}
\end{eqnarray}
The linear transform $c = a + ib$ is represented by the matrix
$\begin{pmatrix} a & -b \\ b & a\end{pmatrix} = r\begin{pmatrix} cos \theta & -sin \theta \\ sin \theta & cos \theta\end{pmatrix}$. 
This is easily recognizable as a planar rotation matrix and is directly used in 
the ``split-layer'' \ref{split-layers} of CoShNet.
Alternatively one may also write complex multiplication (\textsf{cplx-mult}) as amplitude-phase - i.e. $z_1 = a_1e^{i\theta_1}$
and $z_2 = a_2e^{i\theta_2}$:
\begin{equation}
a_1e^{i\theta_1} a_2e^{i\theta_2} = a_1a_2e^{i(\theta_1 + \theta_2)}
\end{equation}
Thus, \textsf{cplx-mult} is equivalent to \textbf{amplitude scaling} and \textbf{phase addition} or shift.
The significance of phase cannot be overstated, since phase encodes shifts in time for wave signals or position in images. The polar form
is valuable for physical intuition, but its direct use in CVnn is less
attractive as Nitta \cite{Nitta2014} shows a polar CVnn is unidentifiable.

\subsection{Efficient complex-linear operator}\label{split-layers}
The proposed CVnn is implemented using ``split-layers''  i.e. to store and propagate real and imaginary parts as two independent real tensors \cite{Trabelsi2017}.
This enables it to leverage existing and highly optimized code of PyTorch
(e.g. real-Autograd) and other repos since most are in $\mathbb{R}$ domain. 
Split-layer use a SoA (Structure-of-Arrays) layout and is critical for efficient SIMD/GPU code.

Since real and imaginary components are propagated separately,
one might be tempted to ask if a CVnn is just a $2 \times$ RVnn.
Hirose's works \cite{Hirose2011}, \cite{Hirose2012} encourage us to look at \textsf{cplx-mult} as
a $2\times 2$ orthogonal matrix or operating in amplitude-phase space as shown in section \ref{cplx-mul}. The real and 
imaginary components might be stored separately, but they are far from just $2$ real channels.
The cross-terms in eq \ref{cplx-mul2} are critical for this intuition (See \cite{Igelnik2001} for more details). The cross-terms allow information to flow
between the real and imaginary part.
This is also the approach taken by SurReal \cite{Chakraborty2020}
which model complex as a product manifold of scaling and planar rotations. 
Complex scaling naturally becomes a group of transitive actions.

\ifArchive
\begin{figure}
\centering
\includegraphics[width=0.31\linewidth]{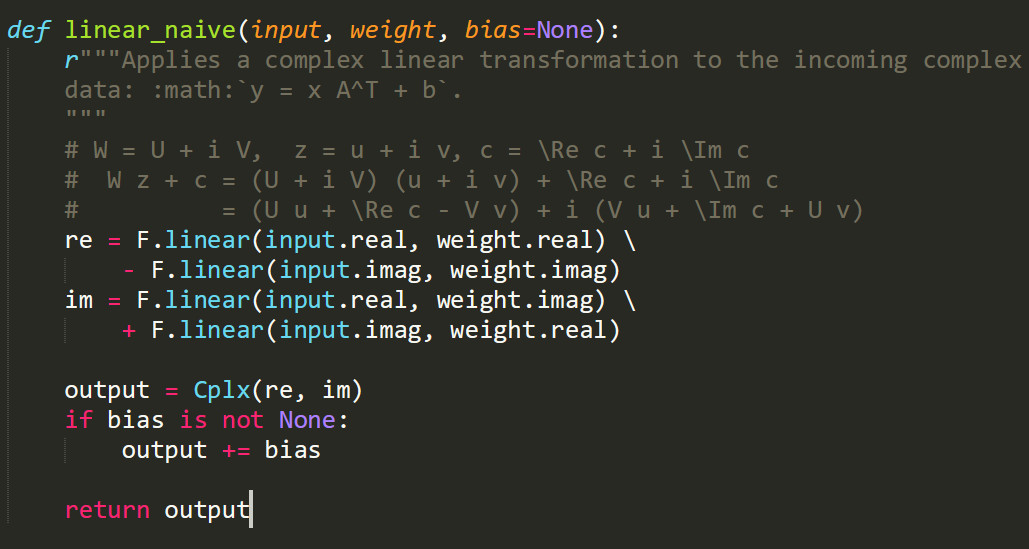}
\hspace{5pt}
\includegraphics[width=0.35\linewidth]{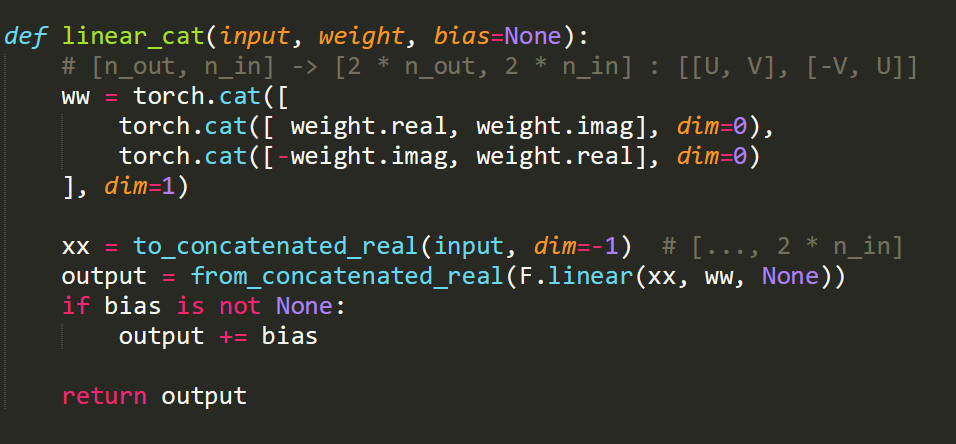}
\caption{\textsf{cplx-module} L: naive \textsf{cplx-linear} R: eq \ref{eqn-3} in code}
\label{fig-hirose-f1}
\end{figure} 
\fi

CoShNet's \textsf{cplx-linear} layers directly use eq.\ref{matrix-cplx-lin-xform} for \textsf{cplx-mult} . 
\ifArchive \footnote{See figure \ref{fig-hirose-f1} in supplement} \fi
This enables the highly optimized autograd of PyTorch to seamlessly
provide gradient support without special coding and saves GPU time.

%
% Efficient cplx-backprop
\subsection{Efficient Differentiation/Backprop in Complex Domain}
\label{cplx-backprop}
The field of complex values is unordered; hence a CVnn is modeled as $f: \mathbb{C} \rightarrow \mathbb{R}$ with a real loss function $\mathcal{L}$. 
Cauchy-Riemann equation
requires a complex differentiable function to be \textit{holomorphic}.
\footnote{holomorphic function are bounded and analytic on the neighborhood of all points in the domain. ReLU is unbounded.
\ifSupplement More in supplement's section 10.2 \fi
}
However, ReLU is non-holomorphic because it is unbounded. 
Let's examine how Wirtinger calculus \cite{Haykin:2002} circumvents the holomorphism requirement.

Similar to the $4\times$ problem of forward pass, during regular backprop
a gradient is just $\dfrac{\partial L(z)}{\partial(z)}$.
For \textsf{cplx-backprop} the direction of steepest ascent
is given by its conjugate-gradient \ref{conjugate-grad} which required 4 partials: 

\begin{equation}
\dfrac{\partial L}{\partial z^*} = \dfrac{\partial L}{\partial u} \times  \dfrac{  \partial u}{\partial z^*} + \dfrac{\partial L}{\partial v} \times \dfrac{\partial v}{\partial z^*}
\label{conjugate-grad}
\end{equation} 

Let us have a look at Wirtinger's complex differentiation. 
Let $f$ be a function of reals $a$ and $b$. We can write $f(z, z^*) = f(a, b)$ where $z = a + ib$ and its complex conjugate $z^* = a - ib$.
Wirtinger's derivatives are:

\vspace{-4pt}
\begin{eqnarray}\label{cogradient}
\frac{\partial f}{\partial \mathbf{z}} &=& 
\frac{1}{2}\left( \frac{\partial f}{\partial a} - i \frac{\partial f}{\partial b}\right) , \\
\label{conjugate-cogradient}
\frac{\partial f}{\partial \mathbf{z^*}} &=& 
\frac{1}{2}\left( \frac{\partial f}{\partial a} + i \frac{\partial f}{\partial b} \right). 
\end{eqnarray}

Eq. \ref{cogradient} gives CVnns the ability to use non-holomorphic activation functions, which are often shown to outperform holomorphic ones. Trabelsi \cite{Trabelsi2017} explored several complex activations and
found the simplest {\small\textsf{split-ReLU}} to be the best. ELU was explored in \cite{Clevert2015} because of its 
negative half-space response (which has interesting properties). We found it to be closely
behind {\small\textsf{split-ReLU}} in experiments. Both equations in \ref{cogradient} are
necessary-and-sufficient, only conjugate-gradient 
($\frac{\partial f}{\partial \mathbf{z^*}}$) is needed using two partials:

\begin{equation}
\dfrac{ds}{dz^*} = \dfrac{dL}{ds^*} \times (\dfrac{ds}{dz})^{*}
\end{equation}

% PyTorch gradient sharing vs cplxmodule
The gradient sharing trick is used by PyTorch. In contrast, \textsf{cplx-linear} directly
use the matrix interpretation in our \ref{split-layers}.
\ifArchive See \ref{fig-hirose-f1} in supplement. \fi
The difference of making this choices can be seen
in the performance difference between CoShNet using \textsf{cplxmodule} (which uses the $2\times 2$ multiplication during forward)
vs. native complex type of PyTorch (which does not use $2 \times 2$ multiplication).
Porting entire CoShNet to PyTorch 1.9 native complex tensors, it
takes $\textbf{0.192}$ and $\textbf{0.355}$ seconds for forward and backprop respectively.
By comparison the \textsf{cplxmodule} based model takes $\textbf{0.099}$ and $\textbf{0.184}$ - i.e. twice as fast. This difference is also partially 
attributed to SoA in our 'split-layer' vs. AoS (Array-of-Structures) used by PyTorch.

\subsection{Complex initialization}\label{cplx-init}
Good initialization is critical
to a CNN. Trabelsi et al. \cite{Trabelsi2017} followed Xavier's \cite{Glorot2010a} reasoning to derive the initialization for complex weights.
Key insight is a complex Gaussian modulus follows a \textit{Rayleigh distribution} 
$\frac{x}{\sigma^2}e^{-x^2/2\sigma^2}$. For weights $W$:

\begin{equation}
 Var(W) = Var(|W|) (\mathbb{E}(|W|)^2)
\end{equation}
We now have a formulation for the variance of $W$ in terms of the variance and expectation of its magnitude, both are analytically computable from the Rayleigh distribution’s single parameter $\sigma$, indicating the \textit{mode}. These are:
\begin{equation}
 \mathbb{E}(|W|) = \sigma \sqrt{\frac{\pi}{2}}, \quad Var(|W|) = \frac{4-\pi}{2}\sigma^2
\end{equation}
Xavier provides a good scaling to keep
the outputs and gradients to be the same order of magnitude. We set $\sigma= \sqrt{\frac{Var[W]}{2}} = 1/\sqrt{n_{in} + n_{out}}$. The magnitude of the complex parameter $W$ is then initialized using the Rayleigh distribution with $\sigma$. The phase is uniformly initialized in range $[-\pi, \pi]$
to make sure phase is covered evenly.

%
% CoShCVnn
%
\section{CoShNet Model} \label{coshcvnn}
\begin{figure}[htbp] %rmv htbp to goto previous config.
\ifECCV	\vspace{-9mm} \fi
	\centering
	\includegraphics[width = 180mm]{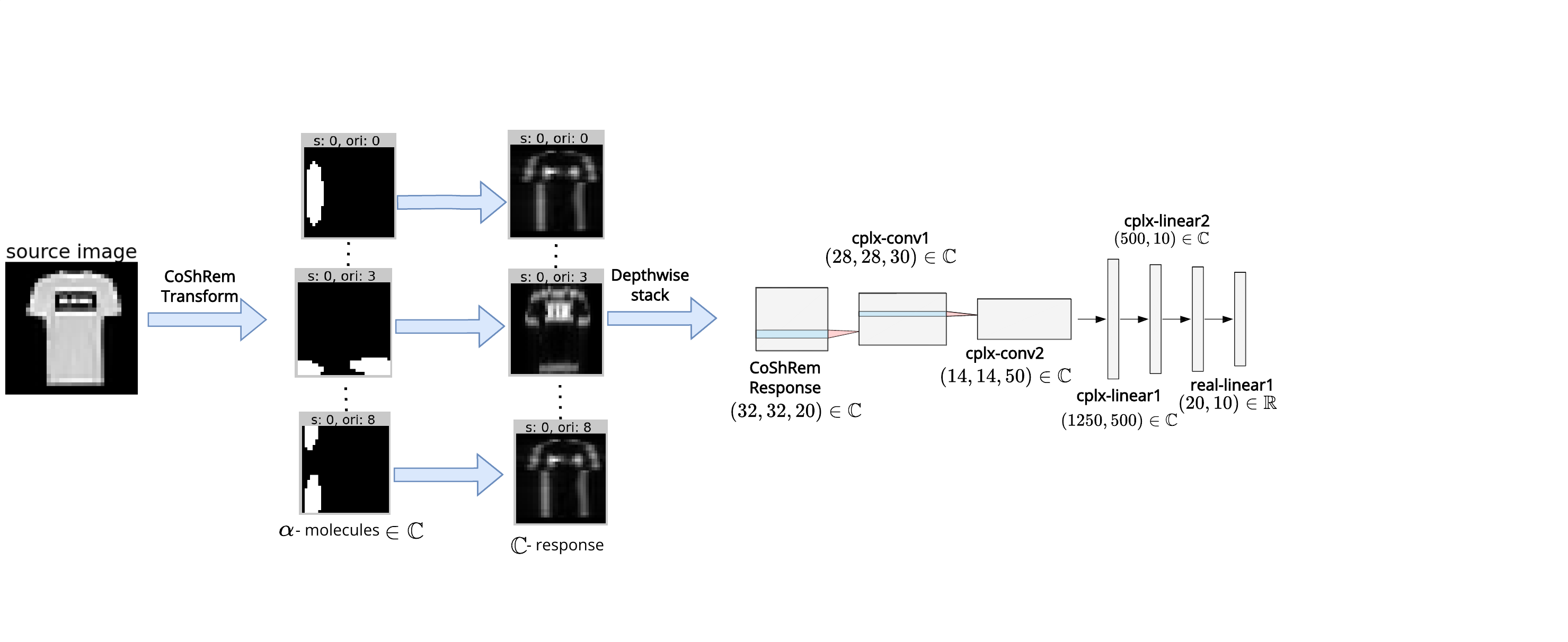}
	\vspace{-4mm}
	\caption{CoShNet with 2 \textsf{cplx-conv} and 3 \textsf{cplx-linear} layers. 
Images are transformed by CoShRem in the data pipeline but acts as a fixed input layer.
Its only uniqueness being its use of complex operators.
	}
	\label{fig:CoShNet}
\ifECCV \vspace{-6mm} \fi
\end{figure}

\noindent
Fig \ref{fig:CoShNet} shows the architecture for CoShNet.
Each of the two convolutional layers has $2\times 2$ average-pooling and
{\small\textsf{split-ReLU}} activation. Following this, the response map is flattened and propagated through $2$
\textsf{cplx-linear} layers. Next, the real and imaginary components of response are
concatenated and passed through an additional linear layer which feed the
cross-entropy loss. CoShNet operates entirely in $\mathbb{C}$ until the last layer.
Especially in the fully connected linear layers to take advantage of the orthogonal
decision boundary of complex layers.
CoShNet is initialized with \textsf{cplx-init} from Trabelsi et al. discussed in \ref{cplx-init}.

%
% what CoShNet does NOT have:
The proposed CVnn is notable for what it does \textbf{not} have:

\ifECCV \vspace{-3pt} \fi
\begin{enumerate}
\item \textit{data-augmentation} is often used to learn a stable embedding.
CoShNet is endowed with a embedding stable to perturbations due to CoShRem's properties.
\item \textit{skip-connections} is usually needed for deep models to pass the higher
resolution signals to upper layers. CoShNet uses a multi-scale transform and naturally
preserve signal resolution.
\item \textit{batch normalization} is needed in deep models to combat vanishing gradients.
CoShNet only needs 5 layers to achieve good performance and do not suffer from vanishing
gradients.
\item \textit{weight-decay} and \textit{dropout regularization} are not needed.
\item \textit{learning rate schedule} and
\ifArchive \footnote{we perform 1 simple learn schedule test just to explore its
impact} \fi
\textit{early stop} are usually used to avoid overfitting. CoShNet avoided overfitting with a fixed transform. In the experiment section one can see CoShNet is very stable to training
regime and initialization \ref{random-seeds2-log}.
\item \textit{hyper-parameter} tuning is important for RVnns and considerable effect
usually goes into it. CoShNet is very stable towards change in learning rate, number of epochs, batch size etc. as shown later in section \ref{experiments}.
\end{enumerate}
CoShNet uses Adam's defaults with a fixed learning rate of $0.001$
and trains on very modest number of epochs (maximum $20$).
It use batch-size of $128$ when training on the 10k set and $256$ when training on 
60k.

This is in stark contrast to a recent paper \cite{wightman2021resnet} ``.. the
\textit{necessity to optimize jointly the architecture and the training procedure}:
..having \textit{the same training procedure is not sufficient} for
comparing the merits of different architectures.'' Which is the opposite of what one
wants to have - a no-fuss, reliable
training procedure for different datasets and models.

%
% Strengths of CoShNet
\subsection{Strengths of CoShNet}
Previous efforts to develop CVnns \cite{Trabelsi2017}, \cite{Nazarov2020}, \cite{Chakraborty2020}
mostly were split-CVnns and ignored the special properties of $\mathbb{C}$ (\ref{cplx-mul}, \ref{split-layers})  or switch from $\mathbb{C}$ to $\mathbb{R}$ in the middle \cite{Brooks2019}. 
Others did not use special initialization \ref{cplx-init} required for complex weights.
Still earlier CVnns spend large effort on finding
good complex holomorphic activation
which were found to be inferior to {\small\textsf{split-ReLU}}.
Others have inefficient back-prop, 
while Trabelsi et al. \cite{Trabelsi2017} require BatchNorm and learn-rate search.

CoShNet is probably the most thorough CVnn, it stays in $\mathbb{C}$ domain especially in the MLP layers \ref{fig:CoShNet}.
Naturally this is inspired by Nitta's decision boundary and critical point \ref{cplx-decision} analysis.
CoShNet is a fully complex CVnn while the better known work by Trabelsi et al. \cite{Trabelsi2017}
and SurReal \cite{Chakraborty2020} are really partial complex/real-CNNs. 
While SurReal treated the merits of a fully complex NN most thoroughly, it
still switch to real after their \textsf{DIST} layer. 
The MLP layers within a CNN performs the actual classification, which means
it is critical for it be endowed with the \textit{orthogonal property} discussed in
S \ref{cplx-decision}. CoShNet use two \textsf{cplx-linear} layers to form its MLP block.

% CoShNet treatment of phase/Hilbert
Another strength of CoShNet in the context of CVnns is how it maps real-valued
images into $\mathbb{C}$ domain. 
In well known CVnns, the imaginary component is set to $0$ while Trabelsi et al. \cite{Trabelsi2017} uses a single residual block to learn the imaginary component. 
Phase recovery is notoriously difficult.
CoShNet use complex even-odd symmetric Gabor-pairs and the Hilbert transform as covered in S. \ref{Hilbert}.
CoShNet uses this principled method from classic signal processing \cite{bracewell1978fourier}.
Instead of treating it as a real to complex conversion, CoShNet uses a spatial to frequency
transform (CoShRem) which encodes the semantic structure of the image in phase.

\ifArchive \subsection{Prior works in CVNN}
CoShNet leveraged these building blocks \cite{Trabelsi2017}, \cite{Nazarov2020}, \cite{Chakraborty2020} for CVNN 
especially the excellent \textsf{cplxmodule}\footnote{https://github.com/ivannz/cplxmodule.git} by Nazarov.
We integrated and adapted all of them into a coherent and efficient framework \textsf{shnetutil} and are releasing it as a building block for other researchers.

\url{https://github.com/Ujjawal-K-Panchal/coshnet}
\fi

%
% Layer Compression (DCF, tt)
%

\section{Compressing CoShNet with DCF, Tensor-Train \& \textit{tiny}}\label{compress}
Although the CoShNet is quite efficient with respect to generalization and training
times. Its 1.3m parameters when compared to a ResNet18 (11.18m parameters) is very good. There
is room for improvement. 
In a CNN the bulk of the computation resides in the convolutional layers while most
of the parameters are in the linear layers. CoShNet attacks it on these two fronts -
adopting DCFNet \cite{Qiu2018} to optimize the 
\textsf{cplx-conv} layers and Tensor-Train \cite{Oseledets2011} for the \textsf{cplx-linear} layers.
The result is a  highly compressed \textit{tiny}-CoShNet. 
It only has $49.9k$ parameters, which is $1/221$ that of ResNet-18, and still achieves SOTA performance on Fashion MNIST.

\subsection{Decomposed Convolutional Filters}\label{DCFNet}
DCFNet \cite{Qiu2018} decomposes convolutional filters into a truncated expansion using a fixed Fourier-Bessel dictionary in the spatial domain (Fig.\ref{fig:dcfnet-f1}).
Only the expansion coefficients are learned from data. Representing the filters in functional bases reduce the number of trainable parameters to those of the expansion coefficients ($KM'\times M$) where $K$ is the number of basis we select from the dictionary.
DCFNet can be seen as dictionary-learning applied to a convolution kernel. 
In addition, the Fourier-Bessel bases captures the low-frequency component which helps to counteract the tendency for CNN to over emphasize fine details, thus reducing overfitting
\footnote{Enrico Mattei recommended DCFNet to the first author in a private conversation}.
Thus, DCFNet approximate the Toeplitz-matrix $W_{\lambda', \lambda}(u)$ of the convolution kernel by
$W_{\lambda', \lambda}(u) = \sum\limits_{k=1}^K (a_{\lambda', \lambda})_k\Psi_k(u)$
using $K$ atoms $\Psi_k$:

\begin{figure}[htbp]
	\vspace{-5mm}
	\centering
	\includegraphics[width=0.55\linewidth]{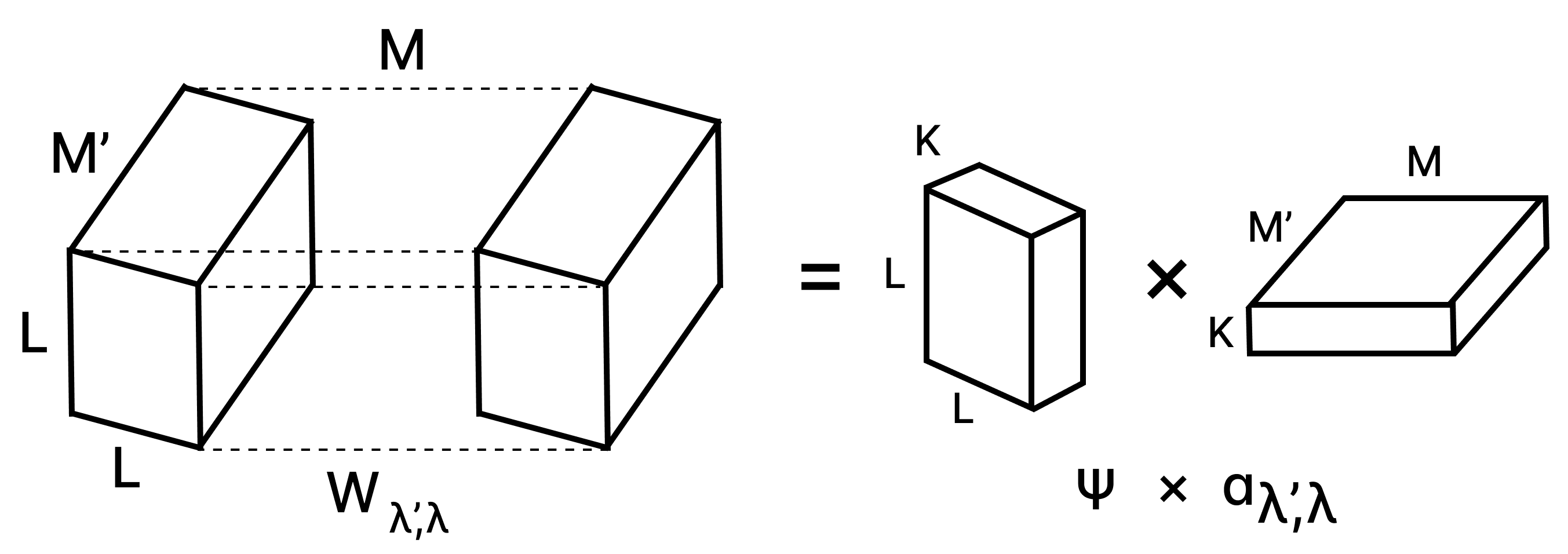}
	\caption{DCFNet: $L\times L\times M' \times M$ conv-layer is decomposed into $K$ atoms of size $L\times L(\Psi)$ and $KM'\times M$ coefficients $a$ where $\Psi$ is fixed and $a$ is learned.}
	\label{fig:dcfnet-f1}
	\vspace{-3mm}
\end{figure}

\noindent
$K=3$ was found to be sufficient for high accuracy.
Hence the savings is $\frac{K}{L^2} \approx \frac{1}{8.3}$ for $L=5$ (for $5\times5$ convolution used in CoShNet). 

\subsection{Tensor Decomposition \& $49.9k$ \textit{tiny}-CoShNet}\label{tiny-CoShNet}
Over $90\%$ of the parameters in CoShNet are in the linear layers. We apply Tensor-Train
factoring to compress them. 
Oseledets \cite{Oseledets2011} introduce the Tensor-Train (TT) factorization to approximate
the  $W \in \mathbb{R}^{M\times N}$ in a linear layer: $f(x) = W \cdot x + B$
using a product factorization:

\begin{equation}\label{TT2}
 \mathcal{A}(j_1, j_2, \ldots, j_d) = \mathbf{G}_1[(j_1)] \mathbf{G}_2[(j_2)]\ldots \mathbf{G}_d[(j_d)]
\approx W
\end{equation}
Tensor-Train (TT) in CoShNet can be seen as a complex version of
TT-\textit{layer} of Novikov et al. \cite{Novikov2015} which introduced
TT to deep learning.

The number of terms in the factorization $\{r_k\}^d_{k=0}$ is the TT-ranks which
controls the approximation quality.
The collections of matrices $\mathbf{G}_k$ are called TT-cores - each is a $3D$ tensor.
All matrices related to the same dim $k$ are restricted to be of size $r_{k-1}\times r_k$.
The rank on the first and last core must be $R_1 = R_{N+1} = 1$ to ensure scalar input and outputs. 
\ifArchive \footnote{the \textit{TT-factoring} in physics is known as Matrix Product States (MPS)}. \fi
The approximation error is controlled by the maximum rank $r_{max}$.
Storage of TT decomposition depends linearly on the number of dimensions
unlike e.g. Tucker which is exponential.
It also has robust approximations algorithms (\textsf{tt-svd} \cite{Oseledets2011})
which we leveraged to produce tt-svd-init \ref{tt-svd-init} covered in the next section.
TT also efficiently supports basic linear algebra operations.

In \textit{tiny}-CoShNet the above two compressions are applied - DCF for the two cplx-conv and TT-\textit{layer} \cite{Novikov2015} for the 
largest of its fully-connected layer. The stunning fact is how small tiny-CoShNet is - $49.9k$ (from $1.3m$).
Converting the dense $1250\mathsf{x}500$ FC-layers to a rank-$8$
TT format ($50\mathsf{x}1\mathsf{x}8\mathsf{x}20$) reduced the parameters by $45\times$
(from $1.25m \mapsto 2.7k$) without sacrificing the expressive power of the layer.
Unlike earlier works which first train an uncompressed model and then apply TT to compress.
The \textit{tiny}-CoShNet is constructed with TT-\textit{layer} \textit{before} training.
This saves considerable time since one does need neither to iterate to compress nor re-train after compression.

%\vspace{-6pt}
\subsubsection{tt-svd-init}\label{tt-svd-init}
Tensor-Train conceptually seems straight forward,
but to implement it well in a CVnn is not so
obvious. First, TT-\textit{layer} \cite{Novikov2015} is implemented with \textsf{cplx-linear}. 
For initialization, at first CoShNet follow other ML researchers and use a simple random
or Xavier init for each tensor-core and the results are not competitive. 
Instead \textit{tiny}-CoShNet first initialize  a dense
\textsf{nn.linear} with \textsf{cplx-init} from Trabelsi et al. \cite{Trabelsi2017}.
Next, \textsf{tt-svd} \cite{Oseledets2011} is used to find the best rank-$k$ TT approximation
for the initialized dense layer.
This ensures the TT-\textit{layer} behaves like a Xavier initialized 
\textsf{cplx-linear} layer at the start of training.
This might seems obvious, but the authors are not aware of previous publications that
approached it this way. Good initialization was found to be key and is a novel
contribution of \textit{tiny}-CoShNet.
This enable \textit{tiny}-CoShNet to use an aggressive compression factor of $45\times$ without
sacrificing performance. 

\ifArchive Supplement details ablation studies where each of the two methods were applied separately. \fi

%
% Experiments.
%
\section{Experiments \& Ablations}\label{experiments}
%All tables, experiments, important ablations along with conclusion should be moved to this section. It should have references to appendix for extra background ablation stuff which donot directly correlate to the central claims made in the previous sections.

To test the hypothesis that CoShNet needs less training data and generalizes 
better most training is performed using the 10k test set of Fashion-MNIST
\cite{FashionMNIST} and 60k train set to test. Additional results using $1k..5k$ to 
train are in table \ref{trainingset-log} .

MNIST is known to be too simple as a dataset, but it is still ubiquitous. 
To make it harder we use the 10k test set to train and $60k$ for test, the top-1 accuracy
after 20 epochs is $98.2$. Additional MNIST results are in
Table \ref{fashion:base-tiny-log}.
All of the remaining experiments use Fashion-MNIST.

Table \ref{fashion:base-tiny-log} compares the performance of the 
already small \textbf{base}-CoShNet ($1.3m$) and the even smaller \textbf{tiny}-CoShNet ($49.9k$).
In the last column of the table, the full training set (60k) is used
to make it easier to compare against other published results.
If we place its performance against the 
FashionMNIST leaderboard, both $92.2$ and $91.4$ place us $8^{th}$ position in the research
papers leaderboard. The $7^{th}$ place use a VGG model with $528m$ parameters. 
A {Resnet-18} based model
\footnote{\href{https://github.com/kefth/fashion-mnist}{kefth}}
with $11.18m$ 
parameters achieved $91.8$ vs. $92.2$ for \textbf{base}-CoShNet.

\begin{table}[htbp]
\ifECCV	\vspace{-4mm} \fi
	\begin{center}
		\begin{tabular}{|l|r|c|c|c|c|}
			\hline \textbf{Model} & \textbf{Dataset} & \textbf{size} & \textbf{epochs} & \textbf{10k training} & \textbf{60k training}\\
			\hline base & Fashion & $1.36m$ & $20$ & $89.2$ (20 epoch) & $92.2$ (20 epoch)\\
			\hline tiny & Fashion & $49.9k$ & $20$ & $88.2$ (18 epoch) & $91.4$ (20 epoch)\\ 
			\hline base & MNIST & $1.36m$ & $20$ & $98.2$ (20 epoch) & $99.1$ (10 epoch)\\
			\hline tiny & MNIST & $49.9k$ & $20$ &  $98.1$ (16 epoch) & $99.2$ (20 epoch)\\ 
			\hline
		\end{tabular}
		\vspace{\tabcapwidth}
		\caption{base-CoShNet vs. \textit{tiny-CoShNet} on Fashion \& MNIST}
		\label{fashion:base-tiny-log}
	\end{center}
\ifECCV	\vspace{-12mm} \fi
\end{table}

\ifSupplement Further ablations and experiments on the two architectures can be found in section $11$ of the supplement. \fi 

\ifArchive
\textit{tiny}-CoShNet Top-1 accuracy is already within $1.0\%$ (88.2 vs 89.2) of the base-model.
Next a simple "learning rate schedule" is performed to see what if any difference it might make.

\subsubsection{\textit{tiny}-CoShNet with learning-rate schedule and over training}\label{TT-DCF-lrschedule}
A simple trial was done with: \textbf{5 epochs at lr=0.01} \& \textbf{15 at lr=0.002}.
The schedule is partially inspired by \href{https://youtu.be/XL07WEc2TRI}{Tishby's 2018 Stanford lecture} in which he described the "two-phases of NN training dynamics" \cite{Shwartz-Ziv2017}. 
From table \ref{TT-DCF-lrschele-log} one can see the tiny $49.9k$ model is neck-to-neck with the baseline using the schedule. The best model was trained in 10 epochs!
The 80 epoch test also shows both models do not degrade with over-training.

\begin{table}[htbp]
	\begin{center}
		\begin{tabular}{|l|r|c|c|}
			\hline Model & size & 20 epochs: 60K best & 80 epochs: 60K best\\
			\hline base & $1366390$ & $\textbf{89.2}$(20E) & $\textbf{89.9}$(77E) \\
			\hline tiny + lr\_schedule & $49990$ & $89.1$(10E) & $89.0$ (10E) \\ 
			\hline
		\end{tabular}
		\vspace{\tabcapwidth}
		\caption{10K20E|10K80E, base-model vs. tiny+lr\_schedule, train=$10k$, test=$60k$, size = number of parameters}
		\label{TT-DCF-lrschele-log}
	\end{center}
\ifECCV	\vspace{-12mm} \fi
\end{table}
\fi

\subsection{CoShNet vs. ResNet-18/50}
This section is dedicated to comparison of CoShNet against the popular ResNet-18 and ResNet-50
in terms of model size, training efficiency, performance, FLOPs etc.

\subsubsection{Model Size \& Convergence}
As shown in following table \ref{tab:coshnetvresnet-fashion}, despite it smaller size (1/8.6), CoShNet-base is consistently better than ResNet-18 in final performance and convergence speed (20 vs. 100 epochs). The \textit{tiny}-CoShNet is even more impressive as it is $1/(11.18*1024/49.99) \approx 1/229$ the size of ResNet-18 but is neck-to-neck with it.

\begin{table}[htbp]
	\centering
	\begin{tabular}{|c|r|c|c|c|c|}
		\hline \textbf{Model} & \textbf{Epochs} & \textbf{\# of Parameters} & \textbf{Size Ratio} & \textbf{Top1 Accuracy (60k)} & \textbf{Accuracy (10k)}\\
		\hline ResNet-18 & 100 & 11.18M     & 229 & 91.8\% & 88.3\%\\
		\hline ResNet-50 & 100 & 23.53M     & 481 & 90.7\% & 87.8\%\\
		\hline CoShNet (base)  & 20 & 1.37M & 28 & \textbf{92.2\%} & \textbf{89.2\%}\\
		\hline \textit{tiny}-CoShNet  & 20  & 49.99K & 1 & 91.6\%  & 88.0\%\\
		\hline
	\end{tabular}
	\vspace{3pt}
	\caption{CoShNet vs. ResNet for Fashion.  CoShNet was only trained for 20 epochs  versus 100 for ResNet.}
	\label{tab:coshnetvresnet-fashion}
\end{table}
Here, 60k means training on the training set (60k) and testing using the 10k test set. Similarly, 10k means training on the test set and testing on training set.

\subsubsection{FLOPs comparison}
In table \ref{FLOPs}, we compare the number of FLOPs per batch for CoShNet and ResNet.

\begin{table}[htbp]
	\centering
	\begin{tabular}{|l|l|r|c|}
		\hline Model  & Batch Size & FLOPs   & FLOP Ratio \\
		\hline ResNet-18 & 128 & 4.77 \textsf{\small GFLOPs}  & 52.48\\
		\hline ResNet-50 & 128 & 10.82 \textsf{\small GFLOPs} & 119\\
		\hline CoShNet (base) & 128 & 93.06 \textsf{\small MFLOPs} & 1\\
		\hline
	\end{tabular}
	\vspace{3pt}
	\caption{FLOPs for CoShNet and ResNet.}
	\label{FLOPs}
\end{table}
CoShNet consume $93$ MFLOPs while ResNet-18 need $4.77$ GFLOPs which is $52$ times more.
ResNet-50 use $119$ times more FLOPs. CVNN usually is considered to be $4\times$ more expensive than their real valued networks. We can see this is not the case for
CoShNet.

\subsection{Initialization, Batchsize and learning-rate stability tests}
Picard \cite{Picard2021}, while studying the influence of random seeds concluded: ``it is surprisingly easy to find an outlier that performs much better or much worse
than the average''. 
On the contrary, Table \ref{random-seeds2-log} shows the impact of random seeds
is negligible  for CoShNet. 
The number of epochs needed to reach the `best model' also is very stable. 
While $4k$ seeds were not tested, the entire model is trained from scratch 
each time with a different seed and not just 1 layer like \cite{Picard2021}.

\ifArchiveEx 
The last column use the ``magic'' seed \footnote{3407 is the ``magic" seed from Picard's 		``\textsf{torch.manual.seed(3407)} is all you need".}
from Picard \cite{Picard2021}.
\fi

%"'Individual differences among deep neural network models" \cite{Mehrer2020}\todo{decide if we will talk about this paper}
\begin{table}[htbp]
\ifECCV \vspace{-4mm} \fi
	\begin{center}
		\begin{tabular}{|l|c|c|c|c|c|c|c|c|c|c|l|}
			\hline \textbf{random seed} & 89.5&89.4&89.3&89.4&89.1&89.6&89.5&89.1&89.7&89.4 & 89.3 \\
			\hline \textbf{batchsize 32..1024} & 89.2&89.4&\textbf{89.5}(128)&89.4&\textbf{89.8}&89.5&89.4&89.6&89.3&88.5&87.2(1024) \\
			\hline \textbf{lr (.001..01)} & 89.3&88.8&88.8&89.0&88.9&88.7&88.7&88.3&88.3&88.5& \\
			\hline
		\end{tabular}
		\vspace{\tabcapwidth}
		\caption{Top-1 accuracy for different random seeds/batch sizes/learning-rate, train $10k$ test $60k$, 20 epochs}
		\label{random-seeds2-log}
	\end{center}
\ifECCV	\vspace{-10mm} \fi
\end{table}

%\subsection{Batchsize tests}
Similar to initialization, batch sizes 
\ifArchive ($32|64|128|160|192|224|256|288|320|512|1024$) \fi %see bsize=128, tiny2-1.log 
and learning rates are evaluated in second and third row of \ref{random-seeds2-log}.
Again the results are very stable.
For scientific integrity, only default initialization and batch size is quoted 
in all the results in this paper. 
One can see some small potential gain if one is willing to do
hyperparameter tuning - e.g. batch size of 192 is $.3\%$ higher. Even then the gain is not dramatic. 
Astute reader will notice CoShNet trained with 10k data and batch of 1024 still
performs remarkably well (87.2). 

\ifArchive
\subsubsection{CoShNet generalization, data efficiency \& RVNN} 
Recent results by Barrachina et al. \cite{Barrachina2021} also confirm our finding that
CVnn do not need drop-out and still generalizes well. Unlike for RVNN
where the authors reported a drop from $92\%$ to $66\%$ without drop-out. 

Additionally, Table \ref{trainingset-log} show using as little as $1k$ to $5k$
training data from Fashion while testing using $60k$,
CoShNet managed an accuracy of around $82..87\%$. 
With 100 samples per class, this is almost in the Few-Shot regime.
In addition, the variability among different subsets of the same size also is small.

\begin{table}[htbp]
 \begin{center}
  \begin{tabular}{|l|c|c|c|c|c|c|c|}
	\hline \textbf{training set}  & 1k & 2k & 3k & 5k & 10k  \\
	\hline \textbf{best model} & ${82.2|81.7|81.8}$ & ${84.7|85.0|84.6}$ & ${85.9|85.3|85.8}$ & ${87.0}|{87.0}|{87.7}$ &  ${89.1}$  \\
	\hline
  \end{tabular}
	\vspace{\tabcapwidth}
 \caption{CoShNet when trained on differently sized subsets of the test set and tested on the 60k training set. The variability among different subsets of the same size is also small in each cell.}
 \label{trainingset-log}
 \end{center}
\ifECCV \vspace{-10mm} \fi
\end{table}
\fi

\ifArchiveEx
\subsubsection{CoShRem Denoise \& Sparsity}

\begin{table}[htbp]
\ifECCV \vspace{-10mm} \fi
 \begin{center}
  \begin{tabular}{|l|c|c|c|}
	\hline threshold sigma (SURE) & .00712 & .01424 & .021359 \\
	\hline Top-1 accuracy & ${89.0}$ & ${88.7}$ & ${88.4}$  \\
	\hline non-zero coefficients & ${70\%}$ & ${58\%}$ & ${50\%}$  \\
	\hline
  \end{tabular}
	\vspace{\tabcapwidth}
 \caption{denoising, 10k train 60k test, 20 epochs}
 \label{denoise-log}
 \end{center}
\ifECCV \vspace{-10mm} \fi
\end{table}
\fi

\ifArchiveEx
\subsection{SurReal based CoShNet}
We use layers from SurReal \cite{Chakraborty2019} based on weighted-Frechet Means 
in our convolutional layers. Very good results can be see in table \ref{fashion:base-tiny-surreal-log}.
\begin{table}[htbp]
	\begin{center}
		\begin{tabular}{|l|r|c|c|c|}
			\hline \textbf{Model} & \textbf{size} & \textbf{epochs} & \textbf{test set} & \textbf{final, best accuracy} \\
			\hline base & $1366390$ & $20$ & $10,000$ & $88.9, \textbf{89.1}$(16E) \\
			\hline tiny & $49990$ & $20$ & $10,000$ & $87.9, \textbf{88.2}$(18E) \\
			\hline surreal & $128219$ & $20$ & $10,000$ & $81.8, 87.2$(13E) \\
			\hline base & $1366390$ & $20$ & $60,000$ &  $\textbf{92.2}, 92.1$(10E) \\
			\hline tiny & $49990$ & $20$ & $60,000$ & $\textbf{91.2}, 90.9$(17E) \\
			\hline surreal & $128219$ & $20$ & $60,000$ & $88.7, 89.7$(17E) \\
			\hline
		\end{tabular}
		\vspace{\tabcapwidth}
		\caption{base-model vs. \textit{tiny} vs. SurReal, Top-1 accuracy}
		\label{fashion:base-tiny-surreal-log}
	\end{center}
\end{table}
\fi

\ifArchive
\subsection{CoShNet under Perturbation \& Noise}
As described previously in section \ref{hybrid-nn}, CoShRem Shearlet transform demonstrate
good robustness to perturbations such as noise and blur. In this section we investigated if this visual robustness translates to better inference and generalization within the CoShNet. The dataset is perturbed with random Gaussian noise ($\mu = 0, \sigma^2 = 10.0$) and random Gaussian blur filter ($\sigma \le 0.75$). CoShNet is trained with different combinations of clean (c) (no perturbation nor noise) and/or perturbed (p) test and/or train sets. The best performance is shown in the following figure \ref{fig:cum-perturb}.

Decent performance in the face of perturbation suggests that the CoShNet is fairly robust to presence of perturbation.

\begin{figure}[htbp]
	\centering
	\includegraphics[width = 120mm]{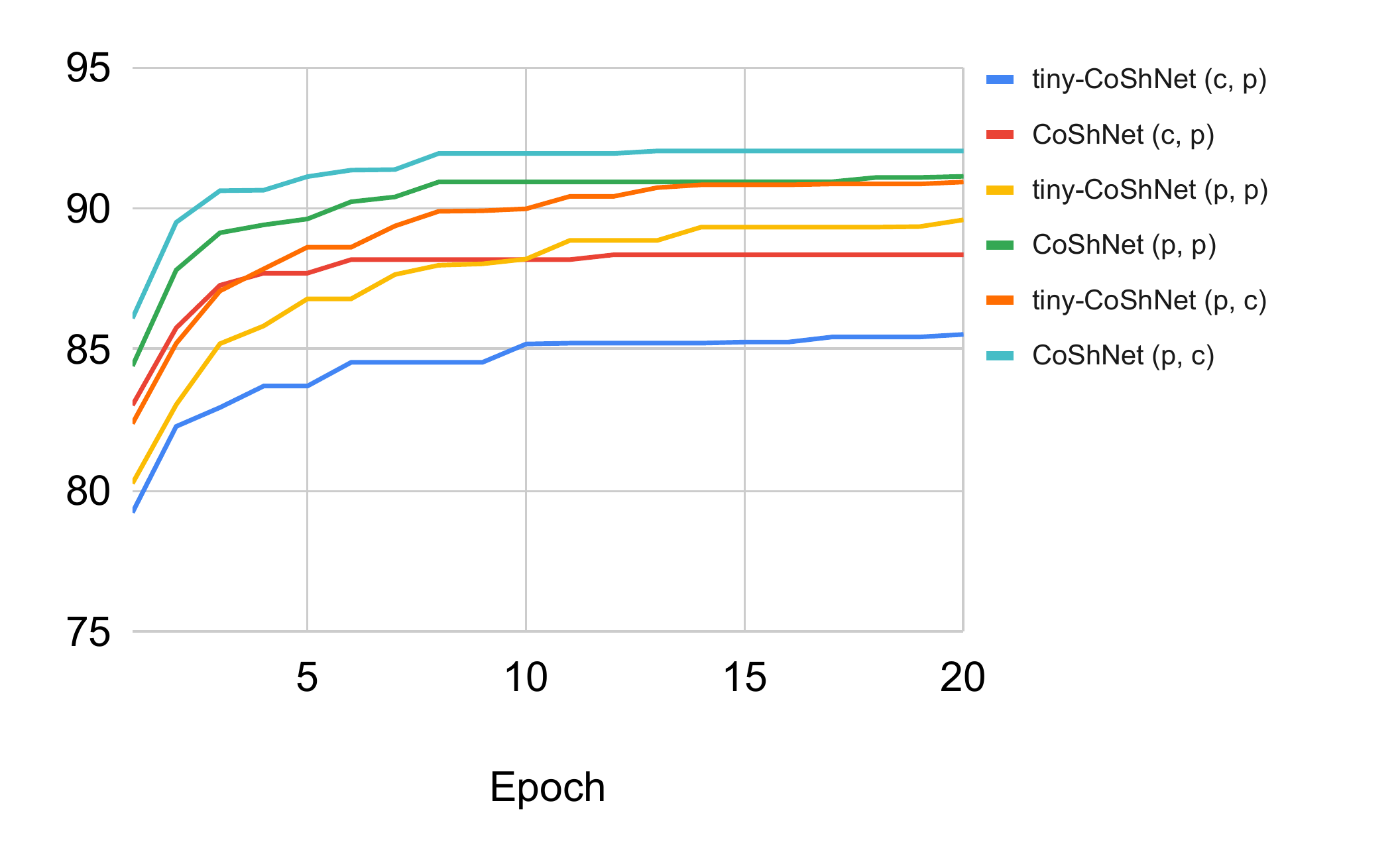}
	\caption{Robustness in presence of perturbation and noise. In legend, clean is abbreviated as 'c', perturbed is abbreviated as 'p'. CoShNet(c, p) means it is trained on a clean training set and then tested on a perturbed test set. Likewise, all other combinations for CoShNet and tiny-CoShNet are trained for 20 epochs and best test-accuracy is reported.}
	\label{fig:cum-perturb}
\end{figure}

\fi

%
% conclusion
%
\ifECCV \vspace{-5mm} \fi
\section{Conclusion}
% In regards with the central claims made in the paper and later proven/disproven by experiments, make strong conclusion which reflects the key takeaways of the paper.
This article introduced a novel hybrid-CVnn architecture using complex-valued shearlet transform.
It is more than $200\times$ smaller than ResNet-18, use $100\times$ less FLOPs and train faster than their real valued counterparts for similar performance. It generalizes extremely well and trains with little labeled data. 
CoShNet trains fast without hyperparameter tuning, learning rate schedules,
early-stops, batchnorm or regularization etc.

%Some of the most attractive properties like small model size, fast training (convergent in 10 epochs)
%and good generalization (we train using 10k set)
%is consistent with prof. Tishby's seminal work
%\cite{Shwartz-Ziv2017} on "Information Bottleneck" and phase-transition in neural-networks.
%The two distinct phases: 1) empirical error minimization (ERM) and 2) representation compression. The ERM phase
%largely is driven to learn the \textit{input} while gradually receiving feedback from the labels
%via the backprop loss.
%The second phase is the 'diffusion phase' which compresses the
%representation by squeezing out non-informative parts of the representation.
%It is conjectured that the proposed hybrid network is able to short-circuit much of the ERM phase with CoShRem providing 
%rich, informative and stable (deformation) representation without having to learn them.
%The CVNN part also appears to be favorable to the diffusion phase (the complex decision surface, cplx-mul as phase shift).
CVnn have been found to give superior results in naturally complex signals (e.g. MRI \cite{cole2020analysis}, SAR \cite{cole2020analysis}, RF \cite{brown2021charrnets} etc.
\footnote{\cite{cole2020analysis} reported a leap from $87.3\%$ to $99.2\%$ accuracy on wide-angle SAR data with the use of complex features}), but
a recent study \cite{Monning2018} found CVnn to not be competitive for images
or have difficulties with training \cite{Zimmermann2011}.
Experimental results herein challenge us to reassess. A principled conversion of real to complex is critical as well as thoroughly treating all the building blocks
of a fully complex-CNN.

Compelling results using DCF and Tensor-Train in a compressed CVnn (\textit{tiny}-CoShNet) and directly training
the compressed model should be of independent interest.

Besides mobile and IoT applications
\footnote{\href{https://www.tinyml.org/}{tinyML}}
where model size and real-time response is valued, the \textit{tiny}-CoShNet can be
an effective block in other architectures. Such as routing-transformer, twins, network-in-network, 
which requires instantiating many NN within a larger network. 

%
% Appendix
%
%\input{appendix}

%
% Acknowledgements
%
\section{Acknowledgements}
We would like to thank Ivan Nazarov\footnote{the author of \textsf{cplxmodule}}, Ivan Barrientos, Rey Medina, Enrico Mattei, Tohru Nitta, Rudrasis Chakraborty for their valuable advice.

%Acknowledge all the help we have had from: Ivannz, Dr. Frery, Dr. Chakraborty, Enrico and others etc.

\bibliographystyle{ieeetr}
\bibliography{references}

\begin{thebibliography}{10}

\bibitem{JonathanFrankle2019}
M.~C. {Jonathan Frankle}, ``{The Lottery Ticket hypothesis: Finding Sparse,
  Trainable Neural Networks},'' in {\em ICLR '19}, vol.~2, p.~42, 2019.

\bibitem{Oyallon2015}
E.~Oyallon and S.~Mallat, ``{Deep roto-translation scattering for object
  classification},'' in {\em CVPR}, vol.~07-12-June, pp.~2865--2873, IEEE
  Computer Society, oct 2015.

\bibitem{Bruna2012}
J.~Bruna and S.~Mallat, ``{Invariant Scattering Convolution Networks},'' {\em
  PAMI}, vol.~35, pp.~1872--1886, Aug 2013.

\bibitem{Szegedy2013}
C.~Szegedy, W.~Zaremba, I.~Sutskever, J.~Bruna, D.~Erhan, I.~Goodfellow, and
  R.~Fergus, ``{Intriguing properties of neural networks},'' in {\em ICLR '14},
  2014.

\bibitem{Kutyniok2014c}
G.~Kutyniok, W.~Q. Lim, and R.~Reisenhofer, ``{ShearLab 3D: Faithful Digital
  Shearlet Transforms based on Compactly Supported Shearlets},'' {\em ACM Trans
  on Mathematical Software}, vol.~V, no.~212, pp.~1--39, 2014.

\bibitem{Reisenhofer2019}
R.~Reisenhofer and E.~J. King, ``{Edge, Ridge, and Blob Detection with
  Symmetric Molecules},'' {\em SIAM J on Imaging Sciences}, vol.~12, no.~4,
  pp.~1585--1626, 2019.

\bibitem{Kovesi1999}
P.~Kovesi, ``{Image Features from Phase Congruency},'' {\em J of Comp. Vision
  Res.}, vol.~1, p.~27, 1999.

\bibitem{Wiatowski2015}
T.~Wiatowski and H.~B{\"{o}}lcskei, ``{Deep Convolutional Neural Networks Based
  on Semi-Discrete Frames},'' in {\em IEEE Int Symp on Information Theory},
  pp.~1212--1216, 2015.

\bibitem{Nitta2002}
T.~Nitta, ``{On the critical points of the complex-valued neural network},'' in
  {\em ICONIP}, vol.~3, pp.~1099--1103, 2002.

\bibitem{Nitta2013}
T.~Nitta, ``{Local minima in hierarchical structures of complex-valued neural
  networks},'' {\em Neural Networks}, vol.~43, pp.~1--7, 2013.

\bibitem{Nitta2003}
T.~Nitta, ``{Orthogonality of Decision Boundary in Complex-valued Neural
  Networks},'' {\em Neural computation '03}, vol.~16, no.~1, pp.~73--97, 2003.

\bibitem{Qiu2018}
Q.~Qiu, X.~Cheng, R.~Calderbank, and G.~Sapiro, ``{DCFNet: Deep Neural Network
  with Decomposed Convolutional Filters},'' in {\em ICML}, vol.~9,
  pp.~6687--6696, feb 2018.

\bibitem{raghu2017svcca}
M.~Raghu, J.~Gilmer, J.~Yosinski, and J.~Sohl-Dickstein, ``Svcca: Singular
  vector canonical correlation analysis for deep learning dynamics and
  interpretability,'' 2017.

\bibitem{Ravishankar2018}
S.~Ravishankar and B.~Wohlberg, ``{Learning Multi-Layer Transform Models},''
  {\em arXiv:1810.08323}, p.~6, oct 2018.

\bibitem{Grohs2013}
P.~Grohs, S.~Keiper, G.~Kutyniok, and M.~Sch{\"{a}}fer, ``$\alpha$-molecules:
  curvelets, shearlets, ridgelets, and beyond,'' in {\em Wavelets and Sparsity
  XV}, vol.~8858, p.~34, 2013.

\bibitem{Guo2008}
K.~Guo and D.~Labate, ``{Representation of Fourier Integral Operators Using
  Shearlets},'' {\em J of Fourier Analysis and Applications '08}, vol.~14,
  no.~3, pp.~327--371, 2008.

\bibitem{Huang1975}
T.~S. Huang, J.~W. Burnett, and A.~G. Deczky, ``{The Importance of Phase in
  Image Processing Filters},'' {\em ASSP '75}, vol.~23, no.~6, pp.~529--542,
  1975.

\bibitem{Oppenheim1981}
A.~V. Oppenheim and J.~S. Lim, ``{The Importance of Phase in Signals},'' {\em
  Proc of the IEEE '81}, vol.~69, no.~5, pp.~529--541, 1981.

\bibitem{Kovesi2003}
P.~Kovesi, ``{Phase Congruency Detects Corners and Edges},'' in {\em Digital
  Image Computing Techniques and Applications '03}, p.~10, 2003.

\bibitem{Andrade-Loarca2020}
H.~Andrade-Loarca, G.~Kutyniok, and O.~{\"{O}}ktem, ``{Shearlets as feature
  extractor for semantic edge detection: The model-based and data-driven
  realm},'' {\em Proc. Math Phys Eng Sci.}, vol.~476, no.~2243, p.~30, 2020.

\bibitem{Morrone1988}
M.~C. Morrone and D.~C. Burr, ``{Feature detection in human vision: a
  phase-dependent energy model.},'' {\em Proc of the Royal Society of London.
  Series B, Biological Sciences}, vol.~235, no.~1280, pp.~221--245, 1988.

\bibitem{bracewell1978fourier}
R.~Bracewell, {\em The Fourier Transform and its Applications}.
\newblock McGraw-Hill Kogakusha, Ltd., 2nd~ed., 1978.

\bibitem{KovesiMATLABCode}
P.~D. Kovesi, ``{MATLAB} and {Octave} functions for computer vision and image
  processing.''
\newblock Available from: $<$https://www.peterkovesi.com/matlabfns/$>$.

\bibitem{Benvenuto92}
N.~Benvenuto and F.~Piazza, ``On the complex backpropagation algorithm,'' {\em
  IEEE Transactions on Signal Processing}, vol.~40, no.~4, pp.~967--969, 1992.

\bibitem{Nitta1997}
T.~Nitta, ``{An extension of the back-propagation algorithm to complex
  numbers},'' {\em Neural Networks}, vol.~10, no.~8, pp.~1391--1415, 1997.

\bibitem{Hirose2011}
A.~Hirose, ``{Nature of complex number and complex-valued neural networks},''
  in {\em Frontiers of Electrical and Electronic Engineering in China}, 2011.

\bibitem{Nitta2003291}
T.~Nitta, ``{On the inherent property of the decision boundary in
  complex-valued neural networks},'' {\em Neurocomputing}, vol.~50,
  pp.~291--303, 2003.

\bibitem{Fukumizu2000}
K.~Fukumizu and S.-I. Amari, ``{Local minima and plateaus in hierarchical
  structures of multilayer perceptrons},'' {\em Neural Networks '00}, vol.~13,
  no.~3, pp.~317--327, 2000.

\bibitem{lee2016gradient}
J.~D. Lee, M.~Simchowitz, M.~I. Jordan, and B.~Recht, ``Gradient descent
  converges to minimizers,'' 2016.

\bibitem{Jin2017}
C.~Jin, R.~Ge, P.~Netrapalli, S.~M. Kakade, and M.~I. Jordan, ``{How to escape
  saddle points efficiently},'' in {\em ICML '17}, vol.~4, pp.~2727--2752,
  2017.

\bibitem{shi2020}
A.~Shi, ``{Neural Network: How Many Layers and Neurons Are Necessary},'' 2020.

\bibitem{Hirose2009}
A.~Hirose, ``{Complex-valued neural networks: The merits and their origins},''
  in {\em IJCNN}, pp.~1237--1244, 2009.

\bibitem{Nitta2014}
T.~Nitta, ``{Plateau in a polar variable complex-valued neuron},'' in {\em
  ICAART '14}, vol.~1, pp.~526--531, 2014.

\bibitem{Trabelsi2017}
C.~Trabelsi, O.~Bilaniuk, Y.~Zhang, D.~Serdyuk, S.~Subramanian, J.~F. Santos,
  S.~Mehri, N.~Rostamzadeh, Y.~Bengio, and C.~J. Pal, ``{Deep Complex
  Networks},'' in {\em ICLR 2018}, 2017.

\bibitem{Hirose2012}
A.~Hirose and S.~Yoshida, ``{Generalization Characteristics of Complex-Valued
  Feedforward Neural Networks in Relation to Signal Coherence},'' {\em IEEE
  Tran on Neural Networks and Learning Systems}, vol.~23, pp.~541--551, 2012.

\bibitem{Igelnik2001}
B.~Igelnik, M.~Tabib-Azar, and S.~R. Leclair, ``{A net with complex weights},''
  {\em IEEE Trans on Neural Networks}, vol.~12, no.~2, pp.~236--249, 2001.

\bibitem{Chakraborty2020}
R.~Chakraborty, Y.~Xing, and S.~X. Yu, ``{SurReal: Complex-Valued Learning as
  Principled Transformations on a Scaling and Rotation Manifold},'' {\em IEEE
  Trans on Neural Networks and Learning Systems}, pp.~1--12, 2020.

\bibitem{Haykin:2002}
S.~Haykin, {\em Adaptive filter theory}.
\newblock Upper Saddle River, NJ: Prentice Hall, 4th~ed., 2002.

\bibitem{Clevert2015}
D.-A. Clevert, T.~Unterthiner, and S.~Hochreiter, ``{Fast and Accurate Deep
  Network Learning by Exponential Linear Units (ELUs)},'' in {\em ICLR '16},
  pp.~1--13, 2015.

\bibitem{Glorot2010a}
X.~Glorot and Y.~Bengio, ``{Understanding the difficulty of training deep
  feedforward neural networks},'' in {\em AISTATS}, vol.~9, pp.~249--256, 2010.

\bibitem{wightman2021resnet}
R.~Wightman, H.~Touvron, and H.~Jégou, ``Resnet strikes back: An improved
  training procedure in timm,'' 2021.

\bibitem{Nazarov2020}
I.~Nazarov and E.~Burnaev, ``Bayesian sparsification of deep c-valued
  networks,'' in {\em ICML}, pp.~7230--7242, 2020.

\bibitem{Brooks2019}
D.~Brooks, O.~Schwander, F.~Barbaresco, J.~Y. Schneider, and M.~Cord,
  ``{Complex-valued neural networks for fully-temporal micro-Doppler
  classification},'' in {\em Int Radar Sym}, vol.~2019-June, p.~11, 2019.

\bibitem{Oseledets2011}
I.~V. Oseledets, ``{Tensor-train decomposition},'' in {\em SIAM J on Scientific
  Computing}, vol.~33, pp.~2295--2317, 2011.

\bibitem{Novikov2015}
A.~Novikov, D.~Podoprikhin, A.~Osokin, and D.~Vetrov, ``{Tensorizing Neural
  Networks},'' in {\em NIPS}, pp.~1--9, 2015.

\bibitem{FashionMNIST}
H.~Xiao, K.~Rasul, and R.~Vollgraf, ``Fashion-mnist: a novel image dataset for
  benchmarking machine learning algorithms,'' {\em arXiv preprint
  arXiv:1708.07747}, 2017.

\bibitem{Shwartz-Ziv2017}
R.~Shwartz-Ziv and N.~Tishby, ``{Opening the Black Box of Deep Neural Networks
  via Information},'' {\em arXiv:1703.00810}, p.~19, mar 2017.

\bibitem{Picard2021}
D.~Picard, ``{torch.manual seed (3407) is all you need : On the influence of
  random seeds in deep learning architecture for computer vision},'' {\em
  arXiv:2109.08203}, pp.~1--9, sep 2021.

\bibitem{Barrachina2021}
J.~A. Barrachina, C.~Ren, C.~Morisseau, G.~Vieillard, and J.~P. Ovarlez,
  ``{Complex-valued vs. Real-valued neural networks for classification
  perspectives: An example on non-circular data},'' in {\em ICASSP},
  pp.~2990--2994, 2021.

\bibitem{cole2020analysis}
E.~K. Cole, J.~Y. Cheng, J.~M. Pauly, and S.~S. Vasanawala, ``{Analysis of Deep
  Complex-Valued Convolutional Neural Networks for MRI Reconstruction},'' 2020.

\bibitem{brown2021charrnets}
C.~N. Brown, E.~Mattei, and A.~Draganov, ``{ChaRRNets: Channel Robust
  Representation Networks for RF Fingerprinting},'' 2021.

\bibitem{Monning2018}
N.~M{\"{o}}nning and S.~Manandhar, ``{Evaluation of Complex-Valued Neural
  Networks on Real-Valued Classification Tasks},'' {\em arXiv:1811.12351},
  p.~18, 2018.

\bibitem{Zimmermann2011}
H.~G. Zimmermann, A.~Minin, and V.~Kusherbaeva, ``{Comparison of the Complex
  Valued and Real Valued Neural Networks Trained with Gradient Descent and
  Random Search Algorithms},'' in {\em ESANN '11}, pp.~27--29, 2011.

\end{thebibliography}

%
% Supplements
%
\ifSupplement
\input{supplement}
\fi
%
% complex prob logs.
%
%\input{complex_prob}

\end{document}